\documentclass[11pt]{article}

\usepackage[final]{acl}

\usepackage{times}
\usepackage{latexsym}

\usepackage[T1]{fontenc}

\usepackage[utf8]{inputenc}

\usepackage{microtype}

\usepackage{inconsolata}

\usepackage{graphicx}

\usepackage{placeins} 
\usepackage{hyperref}
\usepackage{url}
\usepackage{makecell}
\usepackage{amsmath}
\usepackage{amssymb}
\usepackage{wrapfig} 
\usepackage{caption} 
\usepackage{xcolor}
\usepackage{multirow}
\usepackage{tabularray}
\UseTblrLibrary{booktabs}
\usepackage{CJKutf8}
\usepackage[table]{xcolor}
\usepackage[normalem]{ulem} 
\usepackage{tikz}
\usepackage{arydshln}
\usepackage{tcolorbox}
\tcbuselibrary{breakable}
\usepackage{wrapfig}
\usepackage{cleveref}   
\usepackage{pifont} 
\usepackage{algorithm}
\usepackage{algpseudocode}
\definecolor{myred}{RGB}{255,0,0}
\definecolor{mygreen}{RGB}{0,255,0}
\definecolor{myyellow}{RGB}{150,150,0}
\definecolor{myorange}{RGB}{255,155,53}
\definecolor{false}{RGB}{255,0,0}
\definecolor{true}{RGB}{0,176,80}
\newcommand{\ftMorph}{LLMs also struggle with basic symbol-related tasks such as letter counting \citep{fu2024large,xu-ma-2025-llm} and Chinese character decomposition \citep{wu-etal-2025-impact}, which is beyond our scope.}
\newcommand{\ftChinese}{Although radicals in Chinese characters can hint at pronunciation, LLMs often fail to decompose characters correctly \citep{xu-ma-2025-llm}, so we see little risk of “leakage.”}

\usepackage{subcaption}  

\title{\texttt{Phun-Bench}: Evaluating LLMs on Phonological Understanding in Chinese}

\author{Xing Yue \quad Yongliang Shen \quad  Weiming Lu\thanks{Corresponding author.} \\
  Zhejiang University \\
  \texttt{\{yue\_xing, syl, luwm\}@zju.edu.cn} \\
  }

\begin{document}
\maketitle
\begin{abstract}
Language is a vehicle for thought, intricately tied to sounds, symbols, and meaning. However, most large language model (LLM) research focuses on meaning (semantics) and symbols (spelling) while largely overlooking sounds. Existing benchmarks on LLMs' phonological abilities are either solvable through rote memorization or intertwined with other abilities, making them inadequate to measure LLMs' genuine ability in \textit{phonological understanding}. Here, we present \texttt{Phun-Bench}, a purpose-built Chinese benchmark with diverse tasks and settings across three dimensions (Homophony, Rhyme, and Phonetic Similarity), designed to systematically evaluate LLMs' phonological understanding. Our results show that while LLMs excel at recalling correct pronunciations, they generally struggle to leverage phonological knowledge in the flexible and intuitive way that human speakers do. Moreover, through detailed analyses, we propose a hypothesis regarding the underlying mechanism of LLMs' phonological understanding and ``perception'', highlighting an underexplored frontier for future research. \footnote{We release our dataset and code at \url{https://github.com/xing-stellus-yue/Phun-Bench}.}
\end{abstract}

\section{Introduction}
\label{sec:introduction}
Large language models (LLMs), trained on vast corpora and refined through post-training, have shown impressive performance across domains. With chain-of-thought (CoT, \citealp{wei2022chainofthought}) prompting and reinforcement learning (RL), large reasoning models (LRMs, \citealp{DeepSeek2025DeepSeek,openai2025openai}) can now tackle tasks such as mathematics, coding, QA, and creative writing, often matching or surpassing human-level performance.

From a linguistic perspective, language can be viewed as the integration of three aspects: \emph{sound} (phonology and phonetics), \emph{symbol}\footnote{\ftMorph} (orthography and morphology), and \emph{meaning} (semantics). While the meaning aspect alone suffices for many tasks, there exist practically important scenarios, such as creative writing \citep{ormazabal-etal-2022-poelm}, spelling error correction \citep{liang-etal-2023-disentangled}, and toxicity detection under homophonic cloaking \citep{lu-etal-2023-facilitating,xiao-etal-2024-toxicloakcn}, where a joint understanding of both sound and meaning is essential. For instance, without an understanding of rhyme, we cannot expect a model to generate a well-formed poetic line.

However, LLMs are pre-trained on text corpora and intrinsically lack audio-modal capabilities in input, reasoning, and output. A natural question arises: can LLMs acquire an understanding of sound as an emergent ability? We find that few prior studies have directly evaluated LLMs’ phonological understanding or investigated its underlying mechanism. Existing work has typically focused on specific downstream tasks that rely on phonological ability, such as creative writing (e.g., poetry \citep{ormazabal-etal-2022-poelm}, lyrics \citep{yuan2024rhymeaware}, and puns \citep{chen-etal-2024-u,su2025survey}), humor understanding \citep{he-etal-2025-chumor}, toxicity detection \citep{lu-etal-2023-facilitating,mou2020swe}, and Chinese spelling correction \citep{li-etal-2022-improving-chinese,liang-etal-2023-disentangled}. Yet, in such tasks it is difficult to disentangle the contribution of phonological ability from other capacities, such as semantic knowledge. Some studies instead assess LLMs’ basic phonological knowledge—for instance, grapheme-to-phoneme (G2P, \citealp{cheng2024survey}) and phoneme-to-grapheme (P2G, \citealp{lauc2025polyipa}) conversion, phonological-rule QA \citep{shin2025kobalt}, or rhyme word generation \citep{suvarna-etal-2024-phonologybench}. However, these tasks can often be solved through rote memorization of phonological knowledge (e.g., via online dictionaries) rather than reflecting genuine \textbf{phonological understanding}. For example, an LLM may accurately recall word pronunciations, yet still struggle to judge whether two words sound similar or rhyme, an ability that human speakers acquire intuitively.

We first verify LLMs' extensive Chinese phonological knowledge with a character-to-pinyin test. We then introduce \texttt{Phun-Bench}, a Chinese benchmark purpose-built to evaluate LLMs' phonological understanding across three dimensions (Homophony, Rhyme, and Phonetic Similarity), comprising four tasks: (1) Homophone Recall, (2) Contextual Homophone Recognition, (3) Rhyming Sentence Generation, and (4) Similarity Comparison. Evaluations on LLMs show that while they excel at recalling correct pronunciations, they generally underperform on tasks requiring phonological understanding. Through detailed analyses, we further reveal the challenges LLMs face and the underlying mechanism of their phonological understanding, namely, reasoning via the sound layer. Finally, we evaluate audio models on \texttt{Phun-Bench}, highlighting a promising multimodal direction for advancing phonological tasks.

To summarize, our main contributions are:
\begin{itemize}
 \item (\S~\ref{sec:Phonological Understanding}) We introduce the concept of \emph{phonological understanding}, a missing piece in LLM research, and outline the criteria for its evaluation.
 \item (\S~\ref{sec:phun_bench}) Based on these criteria, we propose \texttt{Phun-Bench}, a purpose-built Chinese benchmark with diverse tasks and settings across three dimensions (Homophony, Rhyme, and Phonetic Similarity), and apply it to assess LLMs' phonological understanding.
 \item (\S~\ref{sec:Experiment}) Through detailed analyses and additional experiments, we reveal the challenges LLMs face and the underlying mechanism of their phonological understanding, offering insights for related fields and future research.
\end{itemize}

\section{Related Work}
\label{sec:RW}
\paragraph{LLMs' Phonological Ability}
Early work on phonology in NLP primarily focuses on grapheme-to-phoneme (G2P) conversion \citep{cheng2024survey}, a key component in text-to-speech (TTS) and automatic speech recognition (ASR). Notable datasets include CMUDict \citep{cmudict}, g2pM \citep{park2020gpm}, and the SIGMORPHON 2020 Shared Task on Multilingual Grapheme-to-Phoneme \citep{gorman-etal-2020-sigmorphon}.
With the emergence of LLMs, some studies benchmark them on elementary language tasks, including morphology, phonology/phonetics and grammar. For example, LMentry \citep{efrat-etal-2023-lmentry} features two phonological tasks (word rhyme judgment and homophone judgment) among other non-phonological tasks, and KoBALT \cite{shin2025kobalt} and KoGEM \cite{kim-etal-2025-polishing} include Korean phonological-rule tasks as part of broader benchmarks. To our knowledge, PhonologyBench \citep{suvarna-etal-2024-phonologybench} is the only benchmark purpose-built to evaluate LLMs' phonological ability in English, comprising three tasks: G2P conversion, syllable counting, and rhyme word generation. These tasks actually test memorization of phonological knowledge. In contrast, \texttt{Phun-Bench} focuses on tasks requiring genuine \emph{phonological understanding}, and we further analyze LLMs’ underlying mechanism of it.

Prior research often focuses on downstream tasks that incidentally require phonological ability, such as \emph{pun recognition}, \emph{rhyme-related generation}. A comprehensive review of these works is provided in Appendix~\ref{app:Additional Related Work}. In particular, HomoP-CN \cite{ma-etal-2025-reasoning} evaluates LLMs’ ability to recover Chinese Internet homophones and distinguishes between reasoning and memorization through perturbation. While our work shares some similarities in isolation design and methodology, we instead probe LLMs’ intrinsic phonological ability directly and systematically.

%

\section{Phonological Understanding}
\label{sec:Phonological Understanding}
\begin{figure*}[!t]
\setlength{\abovecaptionskip}{17pt}
\setlength{\belowcaptionskip}{-5pt}
  \centering
  \hspace*{\fill}%
  \begin{subfigure}{0.53\textwidth}
  
    \centering
    \includegraphics[width=\linewidth]{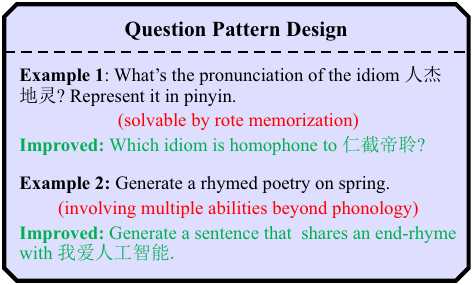}
    \caption{We refine the task design from existing benchmarks.}
    \label{fig:task_design}
  \end{subfigure}%
  \hspace*{\fill}%
  \hspace*{\fill}%
  \begin{subfigure}{0.35\textwidth}
    \small
    \centering
    \resizebox{0.90\linewidth}{!}{%
\begin{tblr}{
    colspec={*{6}{X[c]}},
    width = 0.8\linewidth,
    hline{1,Z} = {1pt,solid},
    hline{2} = {0.5pt,solid},
    hline{3,6,9} = {0.3pt,dashed},
    cell{2}{2,3,4} = {fg=false},
    cell{3}{3,4} = {fg=false},cell{3}{2} = {fg=true},
    cell{4}{2} = {fg=false},cell{4}{3,4} = {fg=true},
    cell{5}{2,3} = {fg=false},cell{5}{4} = {fg=true},
    cell{6}{4} = {fg=false},cell{6}{2,3} = {fg=true},
    cell{7}{3} = {fg=false},cell{7}{2,4} = {fg=true},
    cell{8}{2} = {fg=false},cell{8}{3,4} = {fg=true},
    cell{9}{2,3,4} = {fg=true},
    }
\bf Char. & \bf Init. & \bf Final & \bf Tone & \bf ST & $L_\textrm{sim}$ \\
\begin{CJK}{UTF8}{gbsn}牛\end{CJK} & n & iou & 2 & $T_0$ & 0\\
\begin{CJK}{UTF8}{gbsn}狗\end{CJK} & g & ou & 3 & $T_1$ & 1\\
\begin{CJK}{UTF8}{gbsn}小\end{CJK} & x & iao & 3 & $T_2$ & 1\\
\begin{CJK}{UTF8}{gbsn}书\end{CJK} & sh & u & 1 & $T_3$ & 1\\
\begin{CJK}{UTF8}{gbsn}告\end{CJK} & g & ao & 4 & $T_4$ & 2\\
\begin{CJK}{UTF8}{gbsn}歌\end{CJK} & g & e & 1 & $T_5$ & 2\\
\begin{CJK}{UTF8}{gbsn}猫\end{CJK} & m & ao & 1 & $T_6$ & 2\\
\begin{CJK}{UTF8}{gbsn}糕\end{CJK} & g & ao & 1 & $T_7$ & 3\\
\end{tblr}

    }
    \caption{Similarity Types (STs).}
    \label{fig:Similarity Type}
  \end{subfigure}
  \hspace*{\fill}%

  \caption{(a) Our question pattern design compared to those in prior work. (b) Each pinyin is decomposed into \emph{initial}, \emph{final}, and \emph{tone}. We compare these components between two characters to determine the ST. For illustration, each character’s ST is shown relative to \begin{CJK}{UTF8}{gbsn}\textcolor{myorange}{高}\end{CJK} (\emph{gao1}), with \textcolor{true}{green} for match and \textcolor{false}{red} for mismatch.}
  \label{fig:PU_combined}
\end{figure*}

\subsection{Definition}
\label{ssec:Definition}
We propose \emph{phonological understanding}, defined as \uline{the ability to flexibly apply the phonological and phonetic knowledge inherent in the language in novel situations, beyond simple recall}, which serves as a prerequisite for more complex reasoning tasks, such as poetry generation and perturbed toxicity detection. This definition aligns with the distinction between remembering and understanding in Bloom's Taxonomy \citep{bloom1956taxonomy} and between declarative and procedural knowledge proposed by ACT-R \citep{anderson1997actr}, and follows a functional view that understanding can be inferred from task performance \citep{goldstein-stanovsky-2024-zombies}. For instance, native speakers can quickly grasp a homophonic pun, whereas even advanced second-language learners often struggle.

\subsection{How to Evaluate Phonological Understanding?}
\label{ssec:How to evaluate}
While some existing work has investigated LLMs' phonological ability, they often fall short in evaluating LLMs' \emph{phonological understanding}, as illustrated by the examples in Figure~\ref{fig:task_design}.
\begin{itemize}
    \item Some benchmarks target phonological evaluation, but can be solved through rote memorization of phonological knowledge, such as pronunciations or phonological rules.
    \item Other benchmarks focus on specific downstream tasks, such as spelling correction \citep{li-etal-2022-improving-chinese,liang-etal-2023-disentangled}, or poetry and lyrics generation. However, these tasks are not phonology-only in nature and typically rely heavily on semantic ability.
\end{itemize}

So, we propose that evaluating LLMs' genuine phonological understanding should adhere to two criteria:
\begin{enumerate}
 \item \textbf{Memorization-Proof:} Minimize the possibility that LLMs can succeed merely through memorization.
 \item \textbf{Solve with Sounds, not with Meaning:} Exclude the contribution to results from other abilities, such as semantic understanding.
\end{enumerate}

\paragraph{Language Selection}
\label{para:Language Selection}
Additionally, for languages with alphabetic \emph{writing systems} (e.g., English, Greek), the spelling (and thus the token) can itself be phonologically informative (as the results hint in RQ~3 (\S~\ref{sssec:Analysis 3})), making logographic languages like Chinese preferable to reduce such shortcuts\footnote{\ftChinese}. Moreover, Chinese offers a large pool of homophonic characters, each corresponding to exactly one syllable, which further benefits our benchmark design and data curation.

\subsection{Char2pinyin: A Verification Experiment}
\label{ssec:Verification experiment}
Since no existing work evaluated LLMs on Chinese phonological skills, we begin by verifying their
\emph{phonological knowledge}, through the elementary task of \textbf{character-to-pinyin\footnote{Pinyin is the official romanization system adopted in China to represent the pronunciations of characters and words. For an introduction to pinyin, see Appendix~\ref{app:Further}.} (char2pinyin)} conversion, a foundation for phonological understanding. Specifically, we simply prompt LLMs to spell out the pinyin of given characters $c_\textrm{query}$. 

We curate characters from the \emph{List of Commonly-Used Standard Chinese Characters} \citep{ministry2013list} and retain only monophonic characters. This yields 5,345 unique characters as $c_\textrm{query}$. Evaluation on both LLMs and humans show that LLMs generally outperform humans, exhibiting their extensive phonological knowledge. We present the detailed results in Appendix~\ref{aap:verification experiment}.

\section{Phun-Bench}
\label{sec:phun_bench}
\subsection{Benchmark Design}
\label{ssec:Benchmark Design}
We design \texttt{Phun-Bench} following the two Criteria (\S~\ref{ssec:How to evaluate}). Then, for non-audio LLMs, \emph{how can we measure their phonological understanding?} Drawing inspiration from the distributional hypothesis popularized by J.~R.~Firth \citep{firth1957studies}, who proposed that \emph{a word is characterized by the company it keeps}, we argue that \uline{phonological understanding can be indicated by an understanding of the phonological and phonetic \emph{relationships} between word pronunciations}, such as phonetic similarity and rhymes. For example, humans can intuitively judge if two words sound alike and can easily construct a sentence that rhymes with another.


By analyzing the syllabic structure of Chinese, we categorize eight \textbf{similarity types (STs)} between two characters, as presented in Figure~\ref{fig:Similarity Type}, where $T_7$ denotes ``identical'' and $T_0$ denotes ``totally different''. We use $L_\textrm{sim}$ to indicate the similarity level.
Based on this categorization, we design \texttt{Phun-Bench} across three dimensions:
\begin{itemize}
 \item (\S~\ref{ssec:D1}) \textbf{Homophony-Based Understanding:} the ability to recall commonly-used words from their homophones ($T_7$), either in isolation or within context.
 \item (\S~\ref{ssec:D2}) \textbf{Rhyme-Based Understanding:} the ability to generate a sentence that shares a single-rhyme or double-rhyme ($T_{2,\:4,\:6,\:7}$) with a given reference sentence.
 \item (\S~\ref{ssec:D3}) \textbf{Phonetic-Similarity-Based Understanding:} the ability to compare the phonetic similarity between given words ($T_{1\text{--}7}$).
\end{itemize}

These three dimensions are representative rather than exhaustive. We prioritize homophony, rhyme, and phonetic similarity because they recur in downstream tasks (spell checking creative generation, moderation) and map to established constructs: homophony as the basic equivalence relation, phonetic similarity as a graded relation studied in psychoacoustic and embedding-based work \citep{fang2020using}, and rhyme as a core literary device. Task samples of each dimension are presented in Figure~\ref{fig:data sample}, and the statistics are in Appendix~\ref{aap:Statistics}.

\begin{figure*}[!t]
\setlength{\abovecaptionskip}{5pt}
\setlength{\belowcaptionskip}{-5pt}
    \centering
    \includegraphics[width=\linewidth]{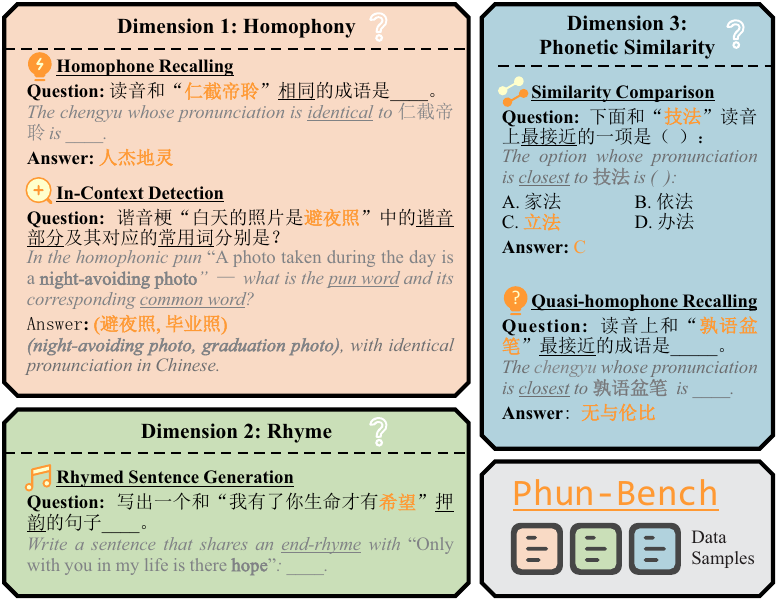}
    \caption{Data samples from \texttt{Phun-Bench}.}
    \label{fig:data sample}
\end{figure*}
\subsection{Dimension 1: Homophony-Based Understanding}
\label{ssec:D1}
 As building connections between words with homophonic relationship is essential to phonological understanding, we evaluate LLMs on recalling words from their homophones, via two tasks. 

\subsubsection{Homophone Recall}
\label{sssec:Homophone Recall}

\paragraph{Task Formulation} Here, we use \emph{chengyu}\footnote{A type of traditional Chinese idiomatic expressions, typically composed of four characters. For an introduction to chengyu, see Appendix~\ref{app:Further}.} as the target word $w_{\textrm{target}}$. Specifically, we provide LLMs with an auto-generated homophone $w_\textrm{ref}$ and prompt them to recover the original chengyu. Since $w_\textrm{ref}$ is probabilistically unlikely to appear in the pre-training corpus and is semantically nonsensical, this approach makes the dataset memorization-proof and eliminates the influence of semantic ability. We use \textbf{exact match (EM)} score to evaluate performance, which indicates if $\hat{w}_\textrm{target}=w_\textrm{target}$.

\paragraph{Dataset Construction} We obtain chengyu from \verb+chinese-idiom-db+\footnote{\url{https://github.com/by-syk/chinese-idiom-db}} and retain only four-character entries. The chengyu are then sorted by frequency in the \verb+news2016zh+\footnote{\url{https://github.com/brightmart/nlp_chinese_corpus}} corpus in decreasing order, and the 2,857 most commonly-used are selected as $w_{\textrm{target}}$ ($freq \geq 10$). To generate $w_\textrm{ref}$, we algorithmically perturb each $w_{\textrm{target}}$ with homophonic commonly-used characters when available (see Algorithm~\ref{alg:homophone}). This results in 1,098 $(w_\textrm{ref},\:w_{\textrm{target}})$ tuples.


\subsubsection{Contextual Homophone Recognition}
\label{sssec:Contextual Homophone Recognition}

\paragraph{Task Formulation} We adapt the homophonic pun detection task for our evaluation. Specifically, LLMs are provided with a pun sentence $s$ and are instructed to (1) detect the \emph{pun word} $w_\textrm{pun}$ within $s$, and (2) recover the corresponding \emph{alternative word} $w_\textrm{target}$. This two-step design enables us to disentangle the contribution of semantic knowledge from phonological understanding. Following \citet{kim2024sure}, we evaluate using the \textbf{cover exact match (CEM)} score for two metrics: \textbf{detection rate (DR)}, i.e., if $\hat{w}_\textrm{pun}$ is covered by $w_\textrm{pun}$, and the \textbf{recovery rate (RR)}, i.e., if $\hat{w}_\textrm{target}$ is covered by $w_\textrm{target}$. 

\paragraph{Dataset Construction} We adapt pun sentences collected by \citet{rohn2024duanzai} as $s$, and then filter out those where $w_\textrm{pun}$ is shorter than three characters or require niche-domain knowledge (e.g., pop culture). Using the \verb+jieba+\footnote{\url{https://github.com/fxsjy/jieba}} and \verb+pypinyin+\footnote{\url{https://github.com/mozillazg/python-pinyin}} toolkits, we convert $w_\textrm{pun}$ and $w_\textrm{target}$ into pinyin $p_\textrm{pun}$ and $p_\textrm{target}$, and manually correct any errors. Only strictly homophonic puns, where $p_\textrm{pun}=p_\textrm{target}$, are retained. This results in 1,684 $(s,\:w_\textrm{pun},\:w_\textrm{target})$ tuples.



\subsection{Dimension 2: Rhyme-Based Understanding}
\label{ssec:D2}
We present the \textbf{Rhyming Sentence Generation (RSG)} task to assess LLMs' rhyme understanding.

\paragraph{Task Formulation} Generating rhyming sentences requires an understanding of the language's phonological system and the relationships between rhyming words. Specifically, we condition LLMs on a reference sentence $s_\textrm{ref}$ and a rhyme rule $\mathcal{R}$, and instruct them to compose a sentence $s_\textrm{gen}$ that rhymes with $s_\textrm{ref}$ (i.e., belong to the same \textbf{rhyme group (RG)} defined in $\mathcal{R}$). To prevent shortcuts via memorizing rhyming words or lyrics, we introduce both a standard \textbf{single-rhyme (1-rhyme)} setting and a more challenging \textbf{double-rhyme (2-rhyme)} setting, in which the last \emph{two} syllables/characters of $s_\textrm{gen}$ must rhyme with those of $s_\textrm{ref}$.


\paragraph{Dataset Construction} We manually define the rhyme rule $\mathcal{R}$, dividing all syllables into 18 RGs ($\mathcal{R} \equiv \{G_i\mid i=1,2,...,18\}$). We then adapt the lyrics from \verb+ChineseLyrics+\footnote{\url{https://github.com/dengxiuqi/ChineseLyrics}} as $s_\textrm{ref}$, filtering out: (1) non-lyric notations (e.g., ``chorus'' or ``songwriter''), (2) sentences containing non-Chinese characters or alphanumerics, and (3) sentences shorter than six characters. Using \verb+jieba+ and \verb+pypinyin+, we convert qualified $s_\textrm{ref}$ to pinyin sequences and obtain their corresponding RGs as $G_\textrm{ref}$. We first randomly sample about 3,000 sentences, then deduplicate and manually filter them to obtain 1,000 $(s_\textrm{ref},\: G_\textrm{ref})$ tuples.

\paragraph{Metrics} Each generate sentence $s_\textrm{gen}$ is converted to pinyin to obtain its RG $G_\textrm{gen}$. We then compute the EM score between $G_\textrm{ref}$ and $G_\textrm{gen}$. Notably, we observe that some models tend to ``plagiarize'' the ending characters from the reference. To prevent this, we explicitly proscribe such behavior in the instruction and disqualify any such generations.

\begin{figure*}[!tb]
\setlength{\belowcaptionskip}{-10pt}
    \centering
    \includegraphics[width=1.0\linewidth]{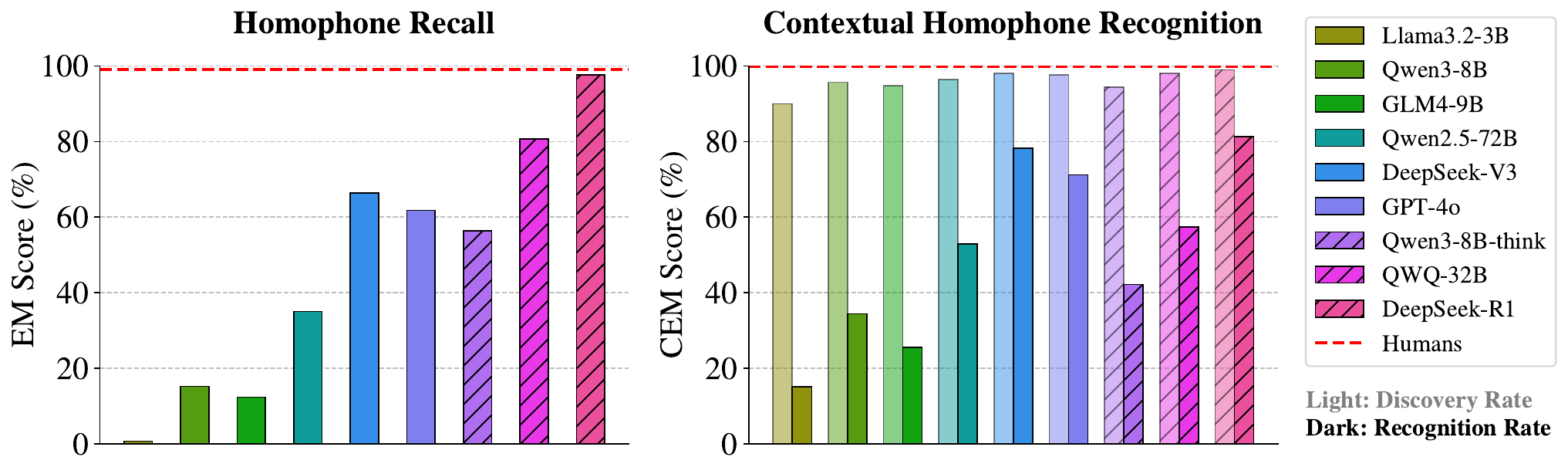}
    \caption{Results of the two homophony-based understanding tasks---HR and CHR. In the right panel, light bars indicate discovery rate (DR) and dark bars indicate recognition rate (RR).}
    \label{fig:D1_res}
\end{figure*}
\subsection{Dimension 3: Phonetic-Similarity-Based Understanding}
\label{ssec:D3}
Building on the proposed STs (\S~\ref{ssec:Benchmark Design}), we introduce the \textbf{Similarity Comparison (SC)} task and generate word perturbations \cite{ma-etal-2025-reasoning} algorithmically to evaluate LLMs’ understanding of phonetic similarity.
\begin{table}[!t]
\setlength{\abovecaptionskip}{5pt}
\setlength{\belowcaptionskip}{-10pt}
\centering
\resizebox{\linewidth}{!}{%
\begin{tblr}{
    colspec={lcccc},
    width = 0.2\linewidth,
    hline{1,Z} = {1pt,solid},
    hline{2} = {0.3pt,solid},
    hline{3} = {0.5pt,solid},
    hline{6} = {0.3pt,dashed},
    }
\SetCell[r=2]{c}\bf PT &\SetCell[r=2]{c}\bf Ans. &\SetCell[r=2]{c}\bf Dist. &\SetCell[c=2]{c} \bf Example & \\
&&& $w_\textrm{ref}$ & $w_\textrm{cand}$ \\
$Pert_{1}$ & $w_{7;\:6}$ & $w_{7;\:1\text{--}3}$ & \begin{CJK}{UTF8}{gbsn}对立\end{CJK}&\begin{CJK}{UTF8}{gbsn} \textcolor{true}{费力} \textcolor{false}{案例} \textcolor{false}{威力} \textcolor{false}{霹雳}\end{CJK} \\
$Pert_{2}$ & $w_{7;\:5}$ & $w_{7;\:1\text{--}3}$ & \begin{CJK}{UTF8}{gbsn}费用\end{CJK}&\begin{CJK}{UTF8}{gbsn} \textcolor{false}{服用} \textcolor{false}{运用} \textcolor{false}{没用} \textcolor{true}{备用}\end{CJK} \\
$Pert_{3}$ & $w_{7;\:4}$ & $w_{7;\:1\text{--}3}$ & \begin{CJK}{UTF8}{gbsn}纪要\end{CJK}&\begin{CJK}{UTF8}{gbsn} \textcolor{false}{将要} \textcolor{false}{需要} \textcolor{true}{机要} \textcolor{false}{概要}\end{CJK} \\
$Pert_{4}$ & $w_{4\text{--}6;\:4\text{--}6}$ & $w_{1\text{--}3;\:1\text{--}3}$ & \begin{CJK}{UTF8}{gbsn}私藏\end{CJK}&\begin{CJK}{UTF8}{gbsn} \textcolor{true}{脂肪} \textcolor{false}{剥离} \textcolor{false}{十足} \textcolor{false}{安装}\end{CJK} \\   
\end{tblr}
}
\caption{Perturbation types (PTs) employed in the SC task. \textcolor{true}{green} denotes answer, and \textcolor{false}{red} denotes distractors.}
\label{tab:sc_perturbation_type}
\end{table}
\paragraph{Task Formulation} We condition LLMs on a reference word $w_\textrm{ref}$ and four candidate words (denoted $w_{\textrm{cand}_i},\:i=1,2,3,4$), and instruct them to select the word (denoted $w_\textrm{close}$) that is \emph{most} phonetically similar to $w_\textrm{ref}$. Both $w_\textrm{ref}$ and each $w_{\textrm{cand}_i}$ are provided either in character form or pinyin form. We use EM score to evaluate performance.

\paragraph{Dataset Construction} We begin with two-character words from \emph{List of Frequently Used Words in Modern Chinese} \citep{state1998list}. These words are sorted by frequency in the \verb+news2016zh+ corpus and the 13,741 most commonly-used words ($freq \geq 100$) are retained. We then apply the \textbf{perturbation types (PTs)} listed in Table~\ref{tab:sc_perturbation_type} to generate candidate words $w_{\textrm{cand}_i}$ (see Algorithm~\ref{alg:similarity}). For example, in the $Pert_1$ setting, we identify two types of words: $w_{7;\:6}$, which contains one $T_7$ character and one $T_{6}$ character, and $w_{7;\:1\text{--}3}$, which contains one $T_7$ character and one $T_{1\text{--}3}$ character, aligned with the characters in $w_\textrm{ref}$. We generate one $w_{7;\:6}$ as the correct answer (denoted $w_\textrm{close}$) and three $w_{7;\:1\text{--}3}$ words as distractors (denoted $w_\textrm{far}$) for each $w_\textrm{ref}$ (when available), which are then shuffled to form the candidate set $w_{\textrm{cand}_i}$.



\section{Experiment}
\label{sec:Experiment}
\subsection{Experiment Settings}
\label{ssec:Experiment Settings}
\paragraph{Evaluated Models} We evaluate the proposed \emph{phonological understanding} (\S~\labelcref{sec:Phonological Understanding,sec:phun_bench}) on eight models with varying sizes and capabilities. \texttt{Qwen3-8B} is evaluated in both modes.
\begin{itemize}
    \item \textbf{Non-Thinking Models:} Llama3.2-3B-Instruct \citep{meta2024llama32}, Qwen3-8B \citep{yang2025qwen3}, GLM4-9B-Chat \citep{zeng2024chatglm}, Qwen2.5-72B-Instruct-AWQ \citep{yang2024qwen25}, DeepSeek-V3 \citep{DeepSeek2024DeepSeek} and GPT-4o \citep{hurst2024gpto}. 
    \item \textbf{Thinking Models:} Qwen3-8B, QwQ-32B-AWQ \citep{qwen2025qwq32b} and DeepSeek-R1 \citep{DeepSeek2025DeepSeek}.
\end{itemize} 

For implementation details, see Appendix~\ref{aap:Model implementation}. Prompts can be found in Appendix~\ref{app:Prompts and Instructions}. Due to budget constraints, we run 300 samples per setting for thinking models, 500 for \texttt{DeepSeek-V3} and \texttt{GPT-4o}, and the full set for the others.

\paragraph{Human Evaluation}
For comparison, we invited three native Chinese undergraduate students to assess \texttt{Phun-Bench}. We randomly sampled 50--100 examples per setting and evenly distributed them among the participants for evaluation. Further details are in Appendix~\ref{aap:Human evaluation}.

\subsection{Dimension 1: Homophony-Based Understanding}
\label{ssec:R_D1}
\subsubsection{Results}
The results in Figure~\ref{fig:D1_res} show that while humans achieve nearly 100\% accuracy, LLMs demonstrate subpar performance in recovering words from their homophones, suggesting that LLMs do not associate homophonic words as naturally as humans. Notably, in the CHR setting, all LLMs achieve a high DR (>90\%) but a substantially lower RR. This indicates that, although LLMs are proficient at detecting pun words using semantic knowledge, they struggle to recover the corresponding alternatives based on phonological understanding.

\subsubsection{RQ 1: How Do LLMs Recover the Homophonic Target Word?}
\label{sssec:Analysis 1}

\begin{figure}[!h]
\setlength{\abovecaptionskip}{5pt}
\setlength{\belowcaptionskip}{-10pt}
    \includegraphics[width=1.0\linewidth]{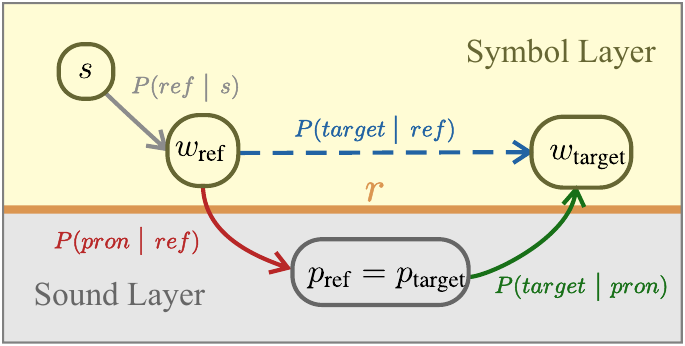}
    \caption{Illustration of the hypothesized mechanism of \emph{phonological understanding via the sound layer}, as in the CHR task.}
    \label{fig:analysis_1}
\end{figure}

We hypothesize that \emph{LLMs perform phonological understanding primarily via pronunciation}, or the \emph{sound layer}, as illustrated in Figure~\ref{fig:analysis_1}. This hypothesis is motivated by distributional asymmetry in pre-training corpora: co-occurrence between a word and its pronunciation is far more frequent than between a word and a perturbation-generated homophone, which we further avoid in our data design. To test this hypothesis, we decompose the \textcolor{azure4}{word2word} task into two steps: \textcolor{red4}{word2pinyin} and \textcolor{green4}{pinyin2word}. Specifically, we prompt models (1) to convert $w_\textrm{ref}$ (from HR) or $w_\textrm{pun}$ (from CHR) into their pinyin $p$ (i.e., \textcolor{red4}{word2pinyin}), and (2) to recover $w_\textrm{target}$ from its pinyin $p$ (i.e., \textcolor{green4}{pinyin2word}). The results are shown in Table~\ref{tab:tab_ana_1}. For HR, we observe that:
{\setlength{\abovedisplayskip}{4pt} \setlength{\belowdisplayskip}{4pt}
\begin{gather}
 \textcolor{red4}{P(pron \mid ref)}\cdot \textcolor{green4}{P(target \mid pron)} \geq \nonumber\\
 \textcolor{azure4}{P(target \mid ref)} \label{eq:eq1}
\end{gather}

And for CHR:
\begin{gather}
 \textcolor{gray6}{P(ref \mid s)} \cdot \textcolor{red4}{P(pron \mid ref)} \cdot \textcolor{green4}{P(target \mid pron)} \geq \nonumber\\
 \textcolor{azure4}{P(target \mid s)} \label{eq:eq2}
\end{gather}
}

These results suggest that LLMs do not effectively associate homophonic words; instead, an indirect two-step conversion via the phonological (sound) layer appears to be easier, thereby supporting our hypothesis empirically. Reasoning traces from thinking models further corroborate this finding (Appendix~\ref{aap:RQ1_Case Study}). A similar investigation on the Character-Level Homophone Recall (CLHR) setting is reported in Appendix~\ref{aap:chr}.
\begin{table*}[th]
\setlength{\abovecaptionskip}{0pt}
\begin{center}


\begin{tblr}{
    colsep=6pt,
    colspec={X[1,c,m]X[2.2,c,m]*{6}{X[r]}X[1.3,r]*{2}{X[r]}X[2.0,c,m]},
    column{3-11}={colsep=4pt},
    rowsep = 0.2pt,
    abovesep = 0pt,
    belowsep=0pt,
    width = \linewidth,
    cells = {font=\footnotesize},
    cell{3,8}{2-11} = {fg=red4},
    cell{4,9}{2-11} = {fg=green4},
    cell{5,10}{2-11} = {fg=azure4},
    cell{6,11}{2-11} = {fg=brown7},
    cell{7}{2-11} = {fg=gray6},
    hline{1,Z} = {1pt,solid},   
    hline{2} = {0.3pt,solid},   
    hline{3} = {0.5pt,solid},   
    hline{7} = {0.3pt,dashed},   
    }

\SetCell[r=2]{m}\bf Task & \SetCell[r=2]{m}\bf Setting &  \SetCell[c=8]{c,m}\bf Model &  &&&&&&&& \SetCell[r=2]{m}\bf Relationship\\
&   & \SetCell[c=1]{c,m} \bf L3.2-3B  & \SetCell[c=1]{c,m} \bf Q3-8B & \SetCell[c=1]{c,m} \bf GLM4-9B  & \SetCell[c=1]{c,m} \bf Q2.5-72B  & \SetCell[c=1]{c,m} \bf DS-V3  & \SetCell[c=1]{c,m} \bf GPT-4o & \SetCell[c=1]{c,m} \bf Q3-8B-think  & \SetCell[c=1]{c,m} \bf QwQ-32B & \SetCell[c=1]{c,m} \bf DS-R1 &   \\ 

\SetCell[r=4]{c}\bf HR  & $P(p \mid w_{q})$ & 0.049 & 0.956 & 0.898 & 0.955 & 1.000 & 0.996 & 0.972 & 0.930 & 1.000 & \SetCell[r=4]{c} 
\(\begin{aligned}
\textcolor{brown7}{r} &= \frac{\textcolor{red4}{P_1} \textcolor{green4}{P_2}}{\textcolor{azure4}{P_3}} \\
  &\ge 1
\end{aligned}
\) \\
& $P(w_{t} \mid p)$ & 0.419 & 0.826 & 0.712 & 0.954 & 0.996 & 0.946 & 0.897 & 0.962 & 1.000 &  \\
 & $P(w_{t} \mid w_{q})$ & 0.007 & 0.152 & 0.124 & 0.351 & 0.664 & 0.618 & 0.563 & 0.807 & 0.977 & \\
 & $r$ & 2.83 & 5.19 & 5.23 & 2.63 & 1.50 & 1.52 & 1.55 & 1.13 & 1.02 &  \\
 
\SetCell[r=5]{c}\bf CHR  & $P(w_{q} \mid s)$ & 0.900 & 0.957 & 0.947 & 0.965 & 0.980 & 0.976 & 0.944 & 0.980 & 0.990 & \SetCell[r=5]{c} 
\(\begin{aligned}
\textcolor{brown7}{r} &= \textcolor{gray6}{P_1} \frac{\textcolor{red4}{P_1} \textcolor{green4}{P_2}}{\textcolor{azure4}{P_3}} \\
  &\ge 1
\end{aligned}
\) \\
 & $P(p \mid w_{q})$ & 0.690 & 0.869 & 0.822 & 0.943 & 0.981 & 0.970 & 0.923 & 0.877 & 0.960 &  \\
 &  $P(w_{t} \mid p)$ & 0.310 & 0.591 & 0.531 & 0.796 & 0.925 & 0.868 & 0.668 & 0.824 & 0.977 &  \\
 & $P(w_{t} \mid s)$ & 0.152 & 0.345 & 0.256 & 0.530 & 0.782 & 0.712 & 0.422 & 0.574 & 0.813 &  \\ 
 & $r$ & 1.27 & 1.43 & 1.61 & 1.37 & 1.14 & 1.15 & 1.38 & 1.23 & 1.14 &  \\ 
\end{tblr}

\end{center}
\caption{Accuracy for \textcolor{red4}{word2pinyin} and \textcolor{green4}{pinyin2word} conversion. Homophone Recall (HR) accuracy is shown as \textcolor{azure4}{$P(w_{t} \mid w_{q})$}, while Contextual Homophone Recognition (CHR) discovery rate (DR) and recognition rate (RR) are represented as \textcolor{gray6}{$P(w_{q} \mid s)$} and \textcolor{azure4}{$P(w_{t} \mid s)$}, respectively. L, Q, and DS denote \texttt{Llama}, \texttt{Qwen} and \texttt{DeepSeek}.}
\label{tab:tab_ana_1}
\end{table*}

\begin{table*}[!bt]
\setlength{\abovecaptionskip}{2pt}
\begin{center}
\resizebox{\linewidth}{!}{%
\small
\begin{tblr}{
colspec={l*{9}{X[c,m]}},
colsep = 3pt,
hline{1,Z}={1pt,solid},
hline{2}={0.5pt,solid},
}
\SetCell{c} \bf Setting & \SetCell{c} \bf L3.2-3B & \SetCell{c} \bf Q3-8B & \SetCell{c} \bf GLM4-9B & \SetCell{c} \bf Q.5-72B & \SetCell{c} \bf DS-V3 & \SetCell{c} \bf GPT-4o & \SetCell{c} \bf Q3-8B-think & \SetCell{c} \bf QwQ-32B & \SetCell{c} \bf DS-R1 \\
\textbf{1-Rhyme} & 0.111 & 0.399 & 0.160 & 0.555 & 0.816 & 0.827 & 0.126 & 0.830 & 0.970 \\
\textbf{2-Rhyme} & 0.014 & 0.051 & 0.009 & 0.053 & 0.083 & 0.072 & 0.022 & 0.437 & 0.848 \\
\end{tblr}

}
\end{center}
\caption{Performance of LLMs on the Rhyming Sentence Generation (RSG) task. L, Q and DS denote \texttt{Llama}, \texttt{Qwen} and \texttt{DeepSeek}, respectively.}
\label{tab:result2}
\end{table*}

\subsection{Dimension 2: Rhyme-Based Understanding}
\label{ssec:R_D2}
\subsubsection{Results}
The results are shown in Table~\ref{tab:result2}. We observe that LLMs generally perform well in 1-rhyme generation but struggle with 2-rhyme generation, showing a clear performance drop. Reasoning appears necessary in this setting, as even humans require more time to compose sentences under complex rhyme constraints.

\subsubsection{RQ 2: What Challenges Do LLMs Face in RSG?}
\label{sssec:Analysis 2}

Analysis of incorrect reasoning traces from \texttt{QwQ-32B} and \texttt{DeepSeek-R1} reveals two main error sources. The first is \emph{indexing error}, where the model fails to correctly identify the last or next-to-last character in either the reference or the generated sentence. The second arises from \emph{misgrouping} characters into incorrect RGs, reflecting a flawed understanding of the Chinese phonological system. For instance, \texttt{DeepSeek-R1} may rhyme \begin{CJK}{UTF8}{gbsn}去 (\emph{qù}, $G_6$)\end{CJK} with \begin{CJK}{UTF8}{gbsn}古 (\emph{gǔ}, $G_7$)\end{CJK} due to their seemingly identical final \emph{u}\footnote{The actual final of \begin{CJK}{UTF8}{gbsn}去\end{CJK} is \emph{ü}, with the diacritic conventionally omitted after the initial \emph{q}.}. In addition, we observe that \texttt{Qwen3-8B-think} is prone to meaningless repetitions of certain sentences or characters, leading to abnormally disastrous performance. Further analyses are provided in Appendix~\ref{app:RQ2}.

\subsection{Dimension 3: Phonetic-Similarity-Based Understanding}
\label{ssec:R_D3}
As mentioned in \S~\ref{ssec:D3}, we use the Similarity Comparison (SC) task to measure LLMs' phonological understanding.

\subsubsection{Results}
\begin{figure}[ht]
\setlength{\abovecaptionskip}{5pt}
\setlength{\belowcaptionskip}{-10pt}
    \centering
    \includegraphics[width=\linewidth]{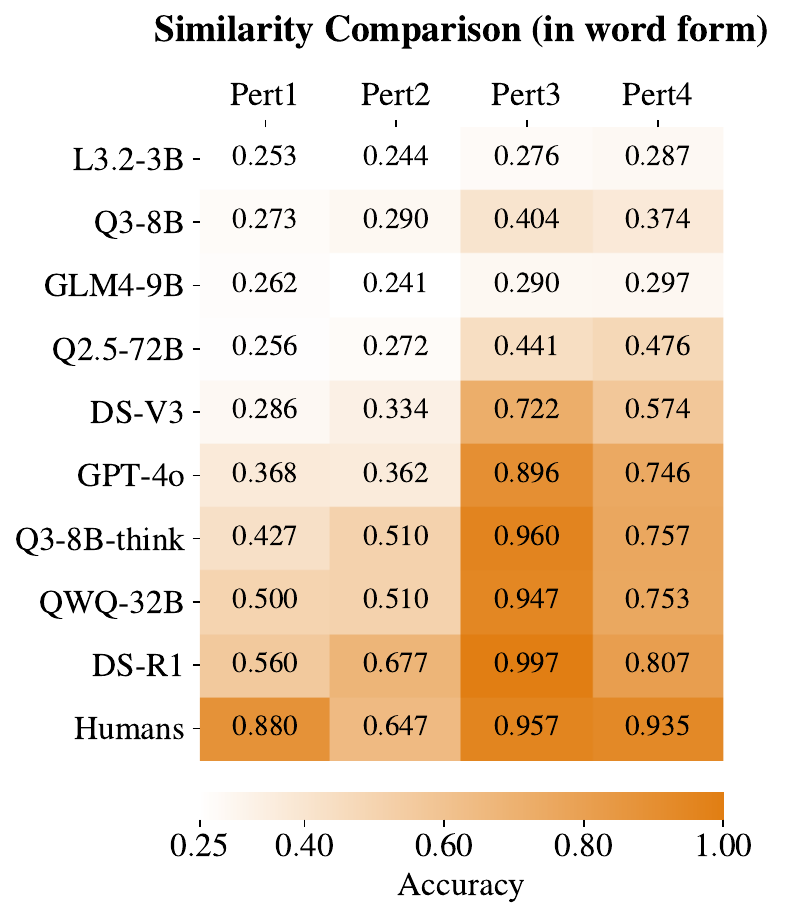}
    \caption{Results of the SC task in word form. L, Q and DS denote \texttt{Llama}, \texttt{Qwen} and \texttt{DeepSeek}, respectively.}
    \label{fig:D3_res_word}
\end{figure}

The results of SC (in word-form setting) are shown in Figure~\ref{fig:D3_res_word}, illustrating that non-thinking models perform poorly, while thinking models achieve some improvement, though still generally below human-level performance (except for the $Pert_3$ setting). Additionally, we provide results for the pinyin-form setting where LLMs show improved performance; detailed results and comparisons are provided in Appendix~\ref{aap:Similarity Comparison in pinyin}.

\subsubsection{RQ 3: How Do LLMs ``Perceive'' Phonetic Similarity?}
\label{sssec:Analysis 3}
While humans typically judge phonetic similarity based on intuition, LLMs appear to rely on reasoning to arrive at their answers, supporting the hypothesis proposed in RQ~1 (\S~\ref{sssec:Analysis 1}). With reasoning, \texttt{Qwen3-8B-think} can even outperform \texttt{DeepSeek-V3} and \texttt{GPT-4o}. By examining the reasoning traces from \texttt{DeepSeek-R1} and \texttt{Qwen3-8B-think} (for details, see Appendix~\ref{aap:RQ3_Case Study}), we observe that they often first convert $w_\textrm{ref}$ and each $w_{\textrm{cand}_i}$ into pinyin, then decompose each syllable into a $(initial, final, tone)$ tuple, and finally compare them using explicit phonological knowledge. Although this ``breaking-down'' reasoning process differs substantially from the intuitive mode of human judgment, it can nevertheless be effective—particularly in $Pert_3$ settings, where $w_\textrm{close}$ and $w_\textrm{ref}$ differ only in tone, a distinction that can be easily captured at the token level. In contrast, changes in the initials (as in $Pert_1$) can ``mislead'' LLMs, leading to poor performance.

\subsubsection{An Exploration: How Do Audio Models Perform?}
\label{sssec:Extended Experiment}
\begin{figure}[!th]
\setlength{\abovecaptionskip}{5pt}
\setlength{\belowcaptionskip}{-12pt}
    \centering
    \includegraphics[width=1.0\linewidth]{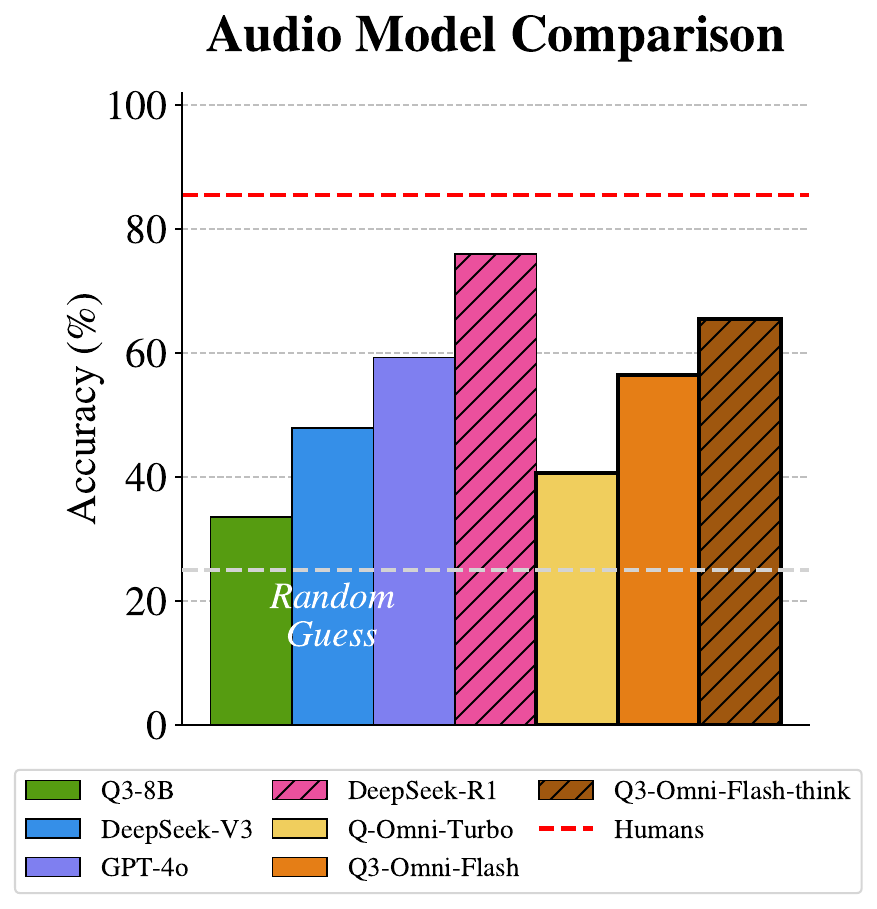}
    \caption{Average Similarity Comparison (SC) results of audio models across four perturbation types (PTs), compared with text-only LLMs.}
    \label{fig:extend_res}
\end{figure}

If LLMs without audio-modal ability must rely on reasoning to compare phonetic similarity, a natural question is whether an audio-enabled LLM would perform better. To investigate, we randomly sample 50 instances from each perturbation type, convert them to audio files using TTS, and evaluate them with \texttt{Qwen-Omni-Turbo}\footnote{Reportedly based on \texttt{Qwen2.5-Omni} \citep{xu2025qwenomni25}.} and \texttt{Qwen3-Omni-Flash \citep{xu2025qwenomni3}}. As shown in Figure~\ref{fig:extend_res}, audio capability improves performance, and with reasoning, \texttt{Qwen3-Omni-Flash} (with $\sim$30B parameters) can even outperform \texttt{GPT-4o} \citep{xu2025qwenomni3}. Further analysis by perturbation type shows differing strengths between text-only and multimodal models (see Appendix~\ref{aap:evaluation on audio LLM}). 

We also inspected thinking traces from \texttt{Qwen3-Omni-Flash}. Many traces first transcribe the audio (sometimes with similar-sounding errors) and then proceed with pinyin-based comparison, often code-switching into English. This suggests current multimodal models may still rely on text-centric reasoning, with audio embeddings underutilized for direct comparison; a deeper analysis on the mechanism of multimodal model's phonological understanding is left for future work.

\subsection{Scaling of Phonological Understanding}
\label{ssec:Qwen3Scaling}
To analyze how LLMs' phonological understanding improves with scale and reasoning, we evaluate the full \texttt{Qwen3} family in both non-thinking and thinking modes on the main tasks. Results in Appendix~\ref{app:Qwen3 Scaling} show consistent gains with model size and a clear boost from thinking mode, while harder tasks (e.g., 2-rhyme RSG) remain challenging even for the largest model.

\section{Discussions}
\label{sec:Discussions}
In summary, our findings suggest that although LLMs possess extensive phonological knowledge, they \uline{generally underperform on tasks requiring phonological understanding}, likely due to limited intrinsic phonological associations between words and difficulty in leveraging phonological knowledge. However, reasoning can elevate performance to a human-comparable level. Based on empirical analysis, we further propose a hypothesis for the underlying mechanism of LLM's phonological understanding, i.e., \uline{reasoning via the sound layer} (\S~\ref{sssec:Analysis 1}), which can be interpreted as a temporary construction of phonological associations. 

Compared to text-only LLMs, the audio model \texttt{Qwen3-Omni-Flash-think} performs better under certain perturbation settings and worse under others (see Figure~\ref{fig:audio_pert_sc}), suggesting a distinct pattern of phonological understanding. However, inspection of reasoning traces indicates that many instances still rely on transcribing audio and reasoning over pinyin, implying a predominantly text-centric pipeline. How multimodality contributes to phonological judgments remains an open question.

Finally, scaling analysis reveals consistent improvements with increasing model size and a clear boost from enabling reasoning, suggesting that phonological understanding benefits from both scale and reasoning. Nevertheless, even the largest model in thinking mode continues to underperform on certain tasks, indicating substantial room for future research.
\section{Conclusion}
\label{sec:Conclusion}
In this work, we introduce the concept of \emph{phonological understanding} and propose criteria for its evaluation. To this end, we present \texttt{Phun-Bench}, a purpose-built Chinese benchmark with various tasks and settings, and assess LLMs across three dimensions: Homophony, Rhyme and Phonetic Similarity. Results show that despite their extensive phonological knowledge, LLMs generally fall short in tasks requiring phonological understanding, while performance scales with model size and is further enhanced by reasoning. We further analyze the \textit{challenges} faced by LLMs and propose a hypothesis regarding the underlying \textit{mechanism} of their phonological understanding, offering insights for related fields and future research. Finally, our exploration of audio models underscores a promising multimodal direction for advancing phonological tasks.

\section*{Acknowledgements}
\label{sec:Acknowledgements}
We appreciate the anonymous reviewers for their valuable comments. This work was supported by the National Natural Science Foundation of China (No.~62376245), National Key Research and Development Project of China (No.~2018AAA0101900), and MOE Engineering Research Center of Digital Library.

\section*{Limitations}
\label{sec:Limitations}
First, our benchmark and analysis focus only on Chinese. While we argue that Chinese is a strong case for evaluating phonological understanding (as discussed in \S~\ref{para:Language Selection}), LLMs may behave differently in other languages. Because each language has its own phonological system and characteristics, directly transferring our benchmark design is generally unsuitable. Extending \texttt{Phun-Bench} to other languages remains an important direction for future work. We acknowledge the potential overgeneralization of the mechanism of phonological understanding we propose. For languages with an alphabetic writing system and more regular grapheme--phoneme mapping (e.g., Italian), explicitly conversion to the sound layer may not be necessary. 

Second, we evaluate only publicly available models (see Appendix~\ref{aap:Model implementation}) without post-training specifically on phonological tasks. Such task-specific fine-tuning could likely improve performance. Nevertheless, we believe that the mechanisms of phonological understanding revealed here would still hold for fine-tuned models.

Finally, our data samples are processed using toolkits and algorithms, followed by manual correction. Although we carefully validated them, minor errors may remain. Similarly, the human baseline we report is based on relatively small samples, and should be viewed as indicative rather than definitive.

\section*{Ethical Considerations}
\label{sec:Ethics Statement}
We manually reviewed the collected pun sentences and lyrics, removing any potentially offensive content, and do not identify significant risks in our work. Human participants were fairly compensated, receiving 80 CNY for approximately 70 minutes of evaluation. We leverage existing artifacts, i.e., toolkits (\verb|jieba|, \verb|pypinyin|) and datasets (\verb|chinese-idiom-db|, \verb|ChineseLyrics|, \verb|news2016zh|, \verb|DuanzAI|), and believe our usage aligns with their intended research purposes.

\bibliography{anthology_part1,anthology_part2,custom}

\appendix
\section{Chinese Language Concepts}
\label{app:Further}
\paragraph{Pinyin}
Pinyin (\begin{CJK}{UTF8}{gbsn}拼音\end{CJK}) is the official romanization system adopted in China, where each syllable consists of an optional \emph{initial} (\begin{CJK}{UTF8}{gbsn}声母\end{CJK}, consonant), a \emph{final} (\begin{CJK}{UTF8}{gbsn}韵母\end{CJK}, vowel with an optional nasal ending) and a diacritic indicating the \emph{tone} (\begin{CJK}{UTF8}{gbsn}声调\end{CJK}). For example, the pinyin of the character \begin{CJK}{UTF8}{gbsn}高\end{CJK} (``tall'') is \emph{gāo}, where \emph{g} is the initial, \emph{ao} is the final, and the macro $\:\bar{}\:$ denotes the first tone. Alternatively, tones can be simply represented numerically, e.g., \emph{gao1}.

\paragraph{Chengyu}
Chengyu (\begin{CJK}{UTF8}{gbsn}成语\end{CJK}) are traditional Chinese idiomatic expressions, typically consisting of four characters, that encapsulate rich cultural meanings and moral lessons in a concise form. They remain an essential part of Chinese language and education, valued for their ability to express complex thoughts elegantly and succinctly. For example, the chengyu \begin{CJK}{UTF8}{gbsn}刻舟求剑\end{CJK} (\emph{kè zhōu qiú jiàn}, literally ``to carve a mark on the boat to seek a lost sword'') means \emph{to act foolishly by sticking to fixed methods despite changing circumstances}. For a comprehensive introduction, see \url{https://en.wikipedia.org/wiki/Chengyu}.

\section{Supplementary Related Work}
\label{app:Additional Related Work}
\subsection{LLMs' Phonological Ability}
We list existing benchmarks that include tasks on LLMs' phonological ability in Table~\ref{tab:comparison}. Some benchmarks also contain tasks targeting other aspects (e.g., morphology, semantics), which we omit here. We further omit monolingual G2P benchmarks, as G2P is an established field with numerous dedicated datasets, and including them would distract from comparison. 

\begin{table*}[t]
\setlength{\abovecaptionskip}{5pt}
\setlength{\belowcaptionskip}{-10pt}
\begin{center}
\begin{tblr}{
colspec={X[2,l]X[1,c]X[2,l]X[1,c]},
hlines={0.3pt,dashed},
hline{1,Z}={1pt,solid},
hline{2}={0.5pt,solid},
width=\linewidth
}
\SetCell[c=1]{c}{\bf Benchmark} & 
\SetCell[c=1]{c}{\bf Year} & 
\SetCell[c=1]{c}{\bf Phonology-Related Tasks} & 
\SetCell[c=1]{c}{\bf Language} \\
ipa-Dict \citep{ipadict} & 2016 & G2P conversion & multilingual \\
g2pM \citep{park2020gpm} & 2020 & polyphone disambiguation & Chinese \\
WikiPron \citep{lee2020massively} & 2020 & G2P conversion & multilingual \\
SIGMORPHON 2020 Shared Task on Multilingual Grapheme-to-Phoneme \citep{gorman-etal-2020-sigmorphon} & 2021 & G2P Conversion & multilingual \\
CharsiuG2P \citep{zhu2022byt} & 2022 & G2P conversion & multilingual \\
LMentry \citep{efrat-etal-2023-lmentry} & 2023 & rhyme word judgment, and homophone judgement & English \\
PhonologyBench \citep{suvarna-etal-2024-phonologybench} & 2024 & G2P conversion, syllable counting and rhyme word generation & English \\
KoBALT \citep{shin2025kobalt} & 2025 & Korean phonological rules & Korean \\
KoGEM \citep{kim-etal-2025-polishing} & 2025 & Korean phonological rules & Korean \\
Multi-LMentry \citep{moroni-etal-2025-multi} & 2025 & rhyme word judgment and homophone judgement & multilingual\\
HomoP-CN \citep{ma-etal-2025-reasoning} & 2025 & Internet homophone recovery & Chinese\\
\end{tblr}
\end{center}
\caption{Overview of existing benchmarks evaluating LLMs' phonological ability. Monolingual G2P datasets are omitted, since they form an established field and would distract from comparison.}
\label{tab:comparison}
\end{table*}

\subsection{Study on Pun}
\label{aap:study on pun}
A pun is a form of humorous wordplay that can be broadly categorized into two types: \emph{homographic puns}, which exploit words that share the same spelling but differ in meaning and pronunciation (e.g., \emph{bass} as a fish vs. a musical instrument), and \emph{homophonic puns}, which use word pairs that sound alike but differ in spelling (e.g. \emph{check} and \emph{Czech}). Prior work mainly focuses on three dimensions: pun generation \citep{su2025survey}, pun detection \citep{zhou-etal-2020-boating,loakman2025comparing}, and pun explanation \citep{xu-etal-2024-good}. Some studies explicitly leverage pronunciation information for recognizing \citep{zhou-etal-2020-boating} or generating \citep{yu-etal-2020-homophonic} homophonic puns.  \citet{chen2023pretrained} construct a Chinese humor dataset and categorize it into three types: harmonic humor (i.e., homophonic puns), ambiguous humor and incongruous humor. Similarly, \citet{rohn2024duanzai} curates a Chinese homophonic pun dataset, \texttt{DuanzAI}. These works primarily focus on developing methods for pun generation and only incidentally touch open LLMs' phonological ability. \emph{We adapt \texttt{DuanzAI} and process it for our Contextual Homophone Recognition (\S~\ref{sssec:Contextual Homophone Recognition}) task to evaluate and analyze LLMs' phonological understanding.}

\subsection{Rhyme-Related Generation}
\label{aap:rhyme-related generation}
Rhyme is a key feature in various literary forms, such as poetry and lyrics. Early systems for automatic poetry generation were largely hand-engineered \citep{gervás2001expert}, and neural approaches learn from collections of real poetry data. For example, \citet{wockener-etal-2021-end} show that recurrent neural networks (RNNs) struggle to capture sound phenomena such as rhyme and alliteration, and CharPoet \citep{yu-etal-2024-charpoet}, a Chinese classical poetry generation system, does not handle the rhyme aspect. Later, models such as PoeLM \citep{ormazabal-etal-2022-poelm} and ByGPT5 \citep{belouadi-eger-2023-bygpt5} were developed to generate verses end-to-end, conditioned on stylistic attributes like rhyme and meter. With the advancement of LLMs, poems generated by GPT-3.5 are even rated more favorably than human-written ones in certain aspects such as rhythm \citep{porter2024aigenerated}. Additionally, \citet{chen-etal-2024-evaluating-diversity} examine diversity in automatic poetry generation and \citet{walsh-etal-2024-sonnet} test whether LLMs can identify the poetic form of given poems. 

Lyrics generation has received relatively little attention \citep{sun-etal-2023-songrewriter}, with most related work instead focusing on music or multimodal song generation \citep{lei2025levo,cideron2024musicrl,ning2025diffrhythm}. Evaluations by \citet{labrak-etal-2025-synthetic} show that LLM-generated lyrics can be difficult for humans to distinguish from human-written ones. 

These studies mainly target poetry and lyrics generation, only tangentially involving rhyme and LLMs' phonological ability. \emph{Building on these insights, we design the Rhyming Sentence Generation task to directly evaluate and analyze LLMs' phonological understanding.}

\section{Phun-Bench Construction Details and Statistics}
\label{app:Bench Statistics}

\subsection{Verification Experiment}
\label{aap:verification experiment}
Considering that there are more than 80k Chinese characters and most of them are rarely used or variants of others, we curate characters from the \emph{List of Commonly-Used Standard Chinese Characters} (\begin{CJK}{UTF8}{gbsn}《通用规范汉字表》\end{CJK}, \citealp{ministry2013list}). The list is published by the government of China, composed of 8,105 characters, categorized into 3 tiers:
\begin{itemize}
    \item \textbf{Tier~1:} including 3,500 characters designated as frequently used characters.
    \item \textbf{Tier~2:} including 3,000 characters designated as commonly-used characters but less frequently used than those in Tier~1.
    \item \textbf{Tier~3:} including 1,605 characters commonly used as names and terminology.
\end{itemize}

Since many characters are polyphones and their prescribed pronunciations can vary across different sources, we simply filter them out with the \verb|pypinyin| toolkit. This yields 5,345 unique characters as $c_\textrm{query}$ for the char2pinyin task.

We then evaluate both LLMs and humans (for experiment setup, see Appendix~\ref{app:Experiment Setup}). The results, presented in Table~\ref{tab:verification experiment}, show that LLMs (with the exception of the smallest 3B model) generally outperform humans, exhibiting their extensive phonological knowledge. Since Tier-3 characters are rarely seen even by native speakers, we therefore retain only Tier-1 and Tier-2 characters in \texttt{Phun-Bench}. 

\begin{table}[htb]
    \centering
    \resizebox{\linewidth}{!}{%
    \begin{tblr}{
      colspec={lrrr},
      width=\linewidth,   
      hline{1,Z}={1pt,solid},
      hline{2}={0.3pt,solid},
      hline{3}={0.5pt,solid},
      hline{9,12}={.3pt,dashed},
      cell{12}{1-4}={bg=gray9},
    }
      \SetCell[r=2]{c,m} \bf Model & \SetCell[c=3]{c,m} \bf Commonness &&\\
      & \SetCell{c,m} \bf Tier-1 & \SetCell{c,m} \bf Tier-2 & \SetCell{c,m} \bf Tier-3 \\
      \texttt{Llama3.2-3B} & 84.02 & 27.87 & 4.48 \\
      \texttt{Qwen3-8B} & 99.29 & 91.18 & 70.83 \\
      \texttt{GLM4-9B} & 99.01 & 79.09 & 25.92 \\
      \texttt{Qwen2.5-72B} & 99.76 & 92.37 & 59.84 \\
      \texttt{DeepSeek-V3} & 100.00 & 99.80 & 82.20 \\
      \texttt{GPT-4o} & 100.00 & 99.00 & 74.80 \\
      \texttt{Qwen3-8B-think} & 99.00 & 91.33 & 66.00 \\
      \texttt{QwQ-32B} & 98.00 & 89.33 & 51.33 \\
      \texttt{DeepSeek-R1} & 100.00 & 99.67 & 83.00 \\
      Humans & 98.00 & 72.00 & 46.00 \\
\end{tblr}
    }
    \caption{Accuracy (\%) of the Verification experiment. We
verify that LLMs exhibit extensive phonological knowledge of the pronunciations of characters.}
    \label{tab:verification experiment}
\end{table}

\subsection{Perturbation Algorithms}
\label{aap:perturbation algorithms}
We present the pseudocode for the perturbation algorithms used in our data construction. 

Algorithm~\ref{alg:homophone} is employed in the Homophone Recall task (\S~\ref{sssec:Homophone Recall}), where the character set $\mathcal{C}$ consists of the 5,345 retained monophonic characters (see \S~\ref{ssec:Verification experiment}), the chengyu set $\mathcal{W}$ consists of the 2,857 retained four-character chengyu (see \S~\ref{sssec:Homophone Recall}), and $\mathcal{M_\textrm{c2p}}$ is implemented using \verb|pypinyin| for individual characters or the pronunciations provided in the original chengyu dataset (and with manual error corrections for any errors). In lines 3--6, we construct the pinyin-to-character map $G$. In lines 7--22, we perturb each four-character chengyu in $\mathcal{W}$ if a homophonic substitution exists for every character. The perturbed result $w_\textrm{ref}$ is provided to LLMs, with the original chengyu $w_\textrm{target}$ serving as the golden answer.

Algorithm~\ref{alg:similarity} is employed in the Similarity Comparison task (\S~\ref{ssec:D3}), where word set $\mathcal{W}$ consists of the retained 13,741 two-character words, and $PT$ denotes one of the perturbation types defined in Table~\ref{tab:sc_perturbation_type}. In lines 7--10, for each reference word $w_\textrm{ref}$, we search for another word $w$ whose two characters match the similarity type specified by $PT$, aligned with the characters in $w_\textrm{ref}$. We then obtain one answer word $w_\textrm{close}$ and three distractor words $w_\textrm{far}$, which are shuffled to construct the four candidate words $w_{\textrm{cand}_i},\:i = 1,2,3,4$.
\begin{algorithm*}
\begin{algorithmic}[1]
\Require Character set $\mathcal{C}$, chengyu set $\mathcal{W}$, character-to-pinyin converter $\mathcal{M}_\textrm{c2p}$
\Ensure Each character $c$ in $\mathcal{C}$ is monophonic
\State Initialize empty pinyin-to-character map $G \gets \{\}$
\State Initialize output list of tuples $\mathcal{T} \gets \varnothing$
\Statex \(\triangleright\) Populate $G$ with each character $c \in \mathcal{C}$
\For{each character $c$ in $\mathcal{C}$}
    \State $p \gets \mathcal{M}_\textrm{c2p}(c)$
    \State $G[p] \gets G[p] \cup \{c\}$   \Comment{$G[p]$ defaults to $\varnothing$ if $p$ is new}
\EndFor

\Statex \(\triangleright\) Perturb each chengyu from $\mathcal{W}$ 
\For{each target chengyu $w_\textrm{target}$ in $\mathcal{W}$}
    \State $w_\textrm{ref\_list} \gets [\_,\:\_,\:\_,\:\_]$  \Comment{Placeholders for 4 characters}
    \For{$i = 0$ to $|w_\textrm{target}|-1$} \Comment{$i=0,1,2,3$}
        \State $c \gets w[i]$
        \State $p \gets \mathcal{M}_\textrm{c2p}(c)$
        \State $c_\textrm{homo} \gets \texttt{sample}(G[p]\setminus \{c\})$
        \If{$c_\textrm{homo} =\texttt{None}$} \Comment{Skip if no available homophonic character}
            \State \textbf{continue}
        \Else
            \State $w_\textrm{ref\_list}[i] \gets c_\textrm{homo}$
            \State $w_\textrm{ref} \gets \texttt{concat}(w_\textrm{ref\_list})$  \Comment{Convert list to string}
        \EndIf
    \EndFor
    \State $\mathcal{T} \gets \mathcal{T} \cup (w_\textrm{ref},\: w_\textrm{target})$
\EndFor
\State \Return $\mathcal{T}$
\end{algorithmic}
\caption{Chengyu Perturbation in the Homophone Recall Task}
\label{alg:homophone}
\end{algorithm*}
\begin{algorithm*}
\begin{algorithmic}[1]
\Require Two-character word set $\mathcal{W}$, perturbation type $PT$
\Ensure $PT \in \{pert_1,\:pert_2,\:pert_3,\:pert_4\}$
\State Initialize output list $\mathcal{T} \gets \varnothing$
\State Extract word-level similarity types for answer and distractors: $WT_\textrm{close},\:WT_\textrm{far} \gets PT_\textrm{pert}$
\State Initialize $w_\textrm{close} \gets \texttt{None}$
\State Initialize $w_\textrm{far} \gets \varnothing$

\For{each reference word $w_\textrm{ref}$ in $\mathcal{W}$}
    \For{each word $w$ in $\mathcal{W}\setminus\{w_\textrm{ref}\}$}
        \Statex \quad \quad \quad \(\triangleright\) {Character-level similarity judgment omitted; see Table~\ref{tab:sc_perturbation_type}}
        \If{$w_\textrm{close} = \texttt{None}$ \textbf{and} $\texttt{get\_type}(w_\textrm{ref},\:w)=WT_\textrm{close}$} 
            \State $w_\textrm{close} \gets w$ \Comment{The answer}
        \ElsIf{$|w_\textrm{far}| < 3$ \textbf{and} $\texttt{get\_type}(w_\textrm{ref},\:w)=WT_\textrm{far}$}
            \State$w_\textrm{far} \gets w_\textrm{far} \cup \{w\}$ \Comment{The distractors}
        \EndIf
        \If{$w_\textrm{close} \neq \texttt{None}$ \textbf{and} $|w_\textrm{far}| \geq 3$}
            \State $w_\textrm{cand},\:i_\textrm{ans} \gets \texttt{shuffle}(\{w_\textrm{close}\} \cup w_\textrm{far})$ \Comment{Shuffle candidates and record the answer index}
            \State $\mathcal{T} \gets \mathcal{T} \cup \{(w_\textrm{ref},\:w_\textrm{cand},\:i_\textrm{ans})\}$
            \State \textbf{break} 
        \EndIf
    \EndFor
\EndFor
\State \Return $\mathcal{T}$
\end{algorithmic}

\caption{Word Perturbation in the Similarity Comparison Task}
\label{alg:similarity}
\end{algorithm*}

\subsection{Statistics}
\label{aap:Statistics}
We present the statistics of \texttt{Phun-Bench} in Table~\ref{tab:statistics}.

\begin{table}[thb]
\setlength{\abovecaptionskip}{5pt}
\setlength{\belowcaptionskip}{-10pt}
\begin{center}
\begin{tblr}{
colspec={X[l]cr},
hlines={0.3pt,dashed},
hline{1,Z}={1pt,solid},
hline{2}={0.5pt,solid},
}
\SetCell[c=1]{c}{\bf Task} & 
\SetCell[c=1]{c}{\bf Setting} & 
\SetCell[c=1]{c}{\bf Size} \\
\SetCell[r=3]{l} char2pinyin & Tier-1 & 2,122\\
& Tier-2 & 2,085\\
& Tier-3 & 1,138\\
\SetCell[r=1]{l} char2char & \SetCell[r=2]{c} - & \SetCell[r=2]{r} 244\\
\SetCell[r=1]{l} pinyin2char & & \\
\SetCell[r=1]{l} Homophone Recall & - & 1,098\\
\SetCell[r=1]{l} Contextual Homophone \newline Recognition & - & 1,684\\
\SetCell[r=2]{l} Rhyming Sentence \newline Generation & 1-rhyme & \SetCell[r=2]{r} 1,000\\
& 2-rhyme & \\
\SetCell[r=4]{l} Similarity Comparison & $pert_1$ & 945\\
 & $pert_2$ & 880\\
 & $pert_3$ & 431\\
 & $pert_4$ & 2,013\\
\end{tblr}
\end{center}
\caption{Dataset size of \texttt{Phun-Bench} and related tasks.}
\label{tab:statistics}
\end{table}

\section{Experiment Setup}
\label{app:Experiment Setup}

\subsection{Model Implementations}
\label{aap:Model implementation}
This section lists the models used in our experiments along with their hyperparameters for reproducibility. When available, \texttt{temperature} and \texttt{top\_p} values follow the official recommendations for better performance.
\begin{itemize}
    \item \textbf{Llama-3.2-3B-Instruct} \citep{meta2024llama32}: open-source, non-thinking. \texttt{temperature} = 0.7, \texttt{top\_p} = 0.8.
    \item \textbf{Qwen3-8B} \citep{yang2025qwen3}: open-source, evaluated in both non-thinking and thinking modes. Non-thinking mode: \texttt{temperature} = 0.7, \texttt{top\_p} = 0.8. Thinking-mode: \texttt{temperature} = 0.6, \texttt{top\_p} = 0.95.
    \item \textbf{GLM-4-9B} \citep{zeng2024chatglm}: open-source, non-thinking. \texttt{temperature} = 0.7, \texttt{top\_p} = 0.8.
    \item \textbf{Qwen2.5-72B-Instruct-AWQ} \citep{yang2024qwen25}: open-source, non-thinking. \texttt{temperature} = 0.7, \texttt{top\_p} = 0.8.
    \item \textbf{QwQ-32B-AWQ} \citep{qwen2025qwq32b}: open-source, thinking. \texttt{temperature} = 0.6, \texttt{top\_p} = 0.95.
    \item \textbf{DeepSeek-V3-0324} \citep{DeepSeek2024DeepSeek}: open-source, non-thinking, accessed via API. \texttt{temperature} = 1.0, \texttt{top\_p} = 0.8.
    \item \textbf{DeepSeek-R1-0528} \citep{DeepSeek2025DeepSeek}: open-source, thinking, accessed via API. \texttt{temperature} = 1.0, \texttt{top\_p} = 0.8.
    \item \textbf{GPT-4o-2024-11-20} \citep{hurst2024gpto}: closed-source, non-thinking, accessed via API. \texttt{temperature} = 1.0, \texttt{top\_p} = 0.8.
    \item \textbf{Qwen-Omni-Turbo-2025-03-26}: multimodal, closed-source, non-thinking, accessed via API. Reportedly based on \texttt{Qwen2.5-Omni}\footnote{\url{https://www.alibabacloud.com/help/en/model-studio/qwen-omni}} \citep{xu2025qwenomni25}. Default hyperparameters are used.
    \item \textbf{Qwen3-Omni-Flash-2025-09-15} \citep{xu2025qwenomni3}: multimodal, closed-source, evaluated in both non-thinking and thinking modes. We access it through API. Default hyperparameters are used.
    \item \textbf{Qwen-TTS-2025-04-10}\footnote{\url{https://www.alibabacloud.com/help/en/model-studio/qwen-tts}}: closed-source, accessed via API. Used to provide audio-modal input for multimodal models.
\end{itemize}

For the scaling analysis on the open-source \texttt{Qwen3} family models in \S~\ref{ssec:Qwen3Scaling}, we access the models via API due to time constraints during the rebuttal phase.

Unless otherwise specified, models are deployed locally via \texttt{Hugging Face}. We use one A6000 GPU for \texttt{Llama-3.2-3B-Instruct}, \texttt{Qwen3-8B}, and \texttt{GLM-4-9B}, and two or four A6000 GPUs for \texttt{Qwen2.5-72B-Instruct-AWQ} and \texttt{QwQ-32B-AWQ}. Inference for non-thinking models takes only minutes per setting. For thinking models, Rhyming Sentence Generation and Similarity Comparison take hours due to thousands of reasoning tokens, while other tasks complete within an hour. As noted in \S~\ref{ssec:Experiment Settings}, budget constraints limit runs to 300 samples per setting for thinking models, 500 for \texttt{DeepSeek-V3} and \texttt{GPT-4o}, and the full set for others. For reproducibility, we fix the random seed at 42.

\subsection{Human Evaluation}
\label{aap:Human evaluation}
All three human participants invited were native Chinese speakers, undergraduate students majoring in clinical medicine, pharmacy, and cultural heritage and museology, respectively, each remunerated with 80 CNY for evaluation. They were not involved in the study design and were informed about data usage before participation. For the char2pinyin verification experiment, we sampled 100 instances from each tier and shuffled them. For the Homophone Recall (HR) and Contextual Homophone Recognition (CHR) tasks, we sampled 100 instances each; for the Similarity Comparison (SC) task, we sampled approximately 50 instances for each perturbation type and shuffled them, across both word-form and pinyin-form settings. The sampled instances were evenly distributed among participants for evaluation, which we estimate took roughly 70 minutes per participant to complete. An instruction is provided for each human-evaluated task (see Appendix~\ref{app:Prompts and Instructions}).

Two participants reported the low readability of the SC task in pinyin form, and the performance was significantly lower, ranging from 2\% to 12\%. Therefore, we only use the word-form results as human baseline.

\subsection{Package Version}
\label{aap:Package Version}
We use the \verb|jieba| (v0.42.1) and \verb|pypinyin| (v0.55.0) packages for Chinese word segmentation and character-to-pinyin conversion.

\section{RQ 1: How do LLMs recover the homophonic target word?}
\label{app:RQ1}

\subsection{Case study}
\label{aap:RQ1_Case Study}
Figure~\labelcref{fig:output_hr_ds,fig:output_hr_q} present two reasoning traces of HR, generated by \texttt{DeepSeek-R1} and \texttt{Qwen3-8B-think}. Notably, the LLMs first convert the homophone into its pronunciation (i.e., pinyin) and then attempt to recover the target word, supporting our two-step hypothesis that LLMs perform phonological understanding via the sound layer.

\subsection{Character-Level Homophone Recall}
\label{aap:chr}
We analyze the mechanism of phonological understanding using the Character-Level Homophone Recall (CLHR) task. Following the \textcolor{red4}{char2pinyin} experiment (\S~\ref{tab:verification experiment}), we use Tier-1 and Tier-2 monophonic characters from the \emph{List of commonly-used Standard Chinese Characters} \citep{ministry2013list} and obtain their pinyin using \verb|pypinyin|. In the \textcolor{azure4}{char2char setting}, LLMs are provided with a character $c_\textrm{ref}$ and prompted to recall five homophonic characters $c_\textrm{target}$; in the \textcolor{green4}{pinyin2char} setting, the pinyin $p$ is provided instead. 

We ensure that for each sample, at least five Tier-1 or Tier-2 monophonic characters are available as answers, resulting in 244 samples. Using \verb|pypinyin|, we check if the five output characters $\hat{c}_\textrm{target}$ have the required pronunciation $p$, assigning a score of 0.2 per correct character (e.g., three correct characters yield 0.6). The average score is then computed across all samples.  

The results are presented in Table~\ref{tab:tab_ana_1_c}. Following the analysis in \S~\ref{sssec:Analysis 1}, we examine the performance relationship between the three tasks and find that the proposed inequality (and hence the hypothesis) does not hold consistently. We attribute this to the fact that LLMs can solve these tasks by memorizing dictionary data, leveraging human-curated associations between homophonic characters, so that an indirect phonological understanding via the sound layer is not strictly necessary.

\begin{table*}[th]
\setlength{\abovecaptionskip}{0pt}
\setlength{\belowcaptionskip}{-5pt}
\begin{center}
\resizebox{\linewidth}{!}{%



\begin{tblr}{
    colsep=6pt,
    colspec={X[1,c,m]X[2.2,c,m]*{6}{X[r]}X[1.3,r]*{2}{X[r]}X[2.0,c,m]},
    column{3-11}={colsep=4pt},
    rowsep = 0pt,
    abovesep = 0pt,
    belowsep=0pt,
    width = \linewidth,
    cells = {font=\footnotesize},
    cell{3}{2-11} = {fg=red4},
    cell{4}{2-11} = {fg=green4},
    cell{5}{2-11} = {fg=azure4},
    cell{6}{2-11} = {fg=brown7},
    hline{1,Z} = {1pt,solid},   
    hline{2} = {0.3pt,solid},   
    hline{3} = {0.5pt,solid},   
    }

\SetCell[r=2]{m}\bf Task & \SetCell[r=2]{m}\bf Setting &  \SetCell[c=8]{c,m}\bf Model &  &&&&&&&& \SetCell[r=2]{m}\bf Relationship\\
&   & \SetCell[c=1]{c,m} \bf L3.2-3B  & \SetCell[c=1]{c,m} \bf Q3-8B & \SetCell[c=1]{c,m} \bf GLM4-9B  & \SetCell[c=1]{c,m} \bf Q2.5-72B  & \SetCell[c=1]{c,m} \bf DS-V3  & \SetCell[c=1]{c,m} \bf GPT-4o & \SetCell[c=1]{c,m} \bf Q3-8B-think  & \SetCell[c=1]{c,m} \bf QwQ-32B & \SetCell[c=1]{c,m} \bf DS-R1 &   \\ 

\SetCell[r=4]{c} \bf CLHR   & $P(p \mid c_{q})$ & 0.161 & 0.809 & 0.523 & 0.760 & 0.999 & 0.995 & 0.952 & 0.937 & 0.998 & \SetCell[r=4]{c}
\(\begin{aligned}
\textcolor{brown7}{r} &= \frac{\textcolor{red4}{P_1} \textcolor{green4}{P_2}}{\textcolor{azure4}{P_3}} \\
  &\ge 1?
\end{aligned}
\) \\
 & $P(c_{t} \mid p)$ & 0.432 & 0.776 & 0.673 & 0.880 & 0.991 & 0.961 & 0.903 & 0.932 & 0.955 &  \\
 &  $P(c_{t} \mid c_{q})$ & 0.125 & 0.420 & 0.388 & 0.668 & 0.949 & 0.917 & 0.778 & 0.840 & 0.959 &  \\
 & $r$ & 0.55 & 1.50 & 0.91 & 1.00 & 1.04 & 1.04 & 1.11 & 1.04 & 0.99 &  \\
\end{tblr}

}
\end{center}
\caption{The accuracy on \textcolor{red4}{char2pinyin}, \textcolor{green4}{pinyin2char} and \textcolor{azure4}{char2char} conversion. L, Q and DS denote \texttt{Llama}, \texttt{Qwen} and \texttt{DeepSeek}, respectively. Note that the inequality doesn't hold consistently.}
\label{tab:tab_ana_1_c}
\end{table*}

\begin{figure*}[bt]
\setlength{\abovecaptionskip}{5pt}
\setlength{\belowcaptionskip}{-5pt}
    \centering
    \includegraphics[width = 0.8\linewidth]{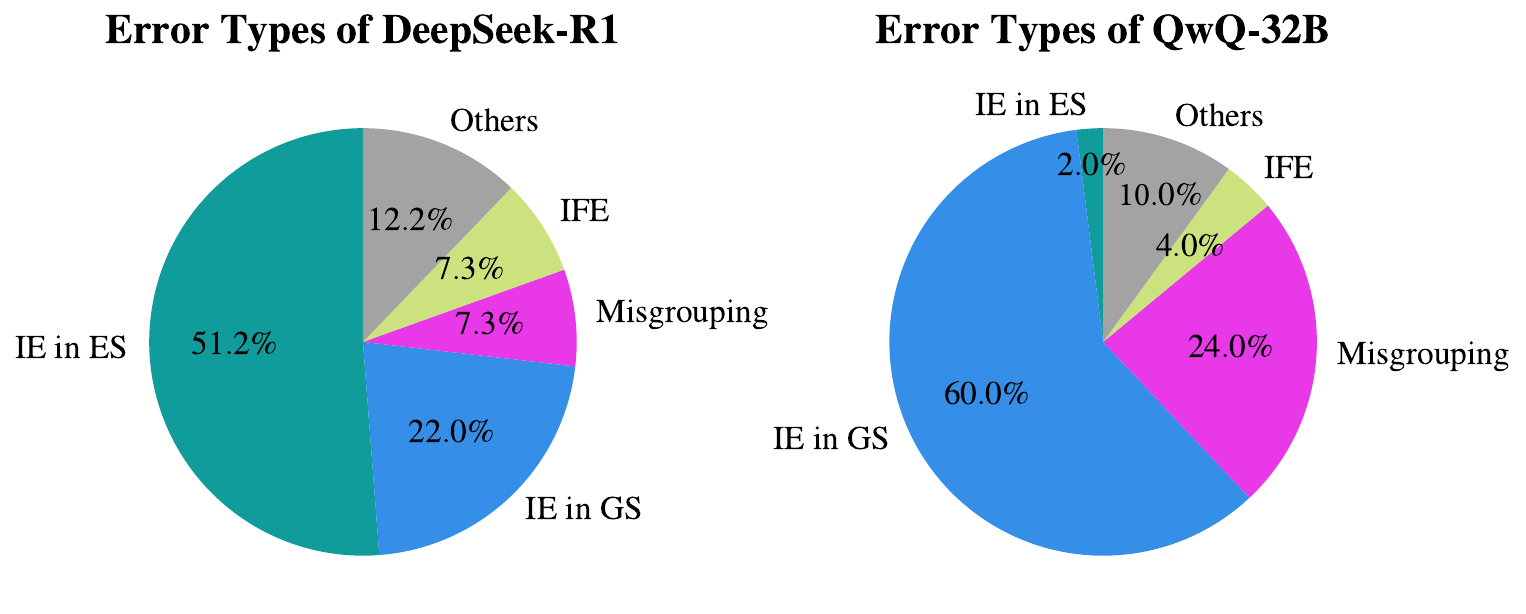}
    \caption{The proportions of error types in the 2-rhyme setting. ES denotes the example sentence; GS denotes the generated sentence.}
    \label{fig:ana2_pie}
\end{figure*}

\section{RQ 2: What Challenges Do LLMs Face in RSG?}
\label{app:RQ2}
We collect 41 failed reasoning traces of the 2-rhyme setting from \texttt{DeepSeek-R1} and another 50 from \texttt{QwQ-32B}, and manually categorize them into four error types. Their proportions are shown in Figure~\ref{fig:ana2_pie}.
\begin{itemize}
    \item \textbf{Indexing Error (IE):} the model fails to correctly identify the last or next-to-last character in either the reference or the generated sentence. For example, rhyming ``\begin{CJK}{UTF8}{gbsn}一直都这样孤独地\underline{\textcolor{red4}{来}}\underline{\textcolor{azure4}{去}}\end{CJK}'' with ``\begin{CJK}{UTF8}{gbsn}独自面对这苍白的衣\underline{\textcolor{red4}{白}}的\underline{\textcolor{azure4}{衣}}\end{CJK}''.
    \item \textbf{Misgrouping:} the model assigns a character to a wrong rhyme group (RG). For example, placing \begin{CJK}{UTF8}{gbsn}飘\end{CJK} (\emph{piāo}) into $G_{11}$ (finals \emph{ou}, \emph{iou}), instead of $G_{10}$ (finals \emph{ao}, \emph{iao}).
    \item \textbf{Instruction-Following Error (IFE):} the model fails to comply with the given instruction. For example, generating a 1-rhyme sentence when a 2-rhyme is required.
    \item \textbf{Others:} miscellaneous errors, such as acknowledging potential rhyme violations but still producing the sentence, or generating a final answer different from the decision in the reasoning trace.
\end{itemize}

For \texttt{Qwen3-8B-think}, we observe an abnormal tendency to produce meaningless repetitions of example sentences, generated sentences, or individual characters/words, across both 1-rhyme and 2-rhyme settings. The model rarely succeeds in genuinely ``composing'' a rhyming sentence; the few correct cases appear more coincidental than deliberate. The underlying cause for this behavior remains unclear.

\section{RQ 3: How Do LLMs ``Perceive'' Phonetic Similarity?}
\label{app:RQ3}

\subsection{Case study}
\label{aap:RQ3_Case Study}
Figure~\labelcref{fig:output_sc_ds,fig:output_sc_q} present two reasoning traces of HR, generated by \texttt{DeepSeek-R1} and \texttt{Qwen3-8B-think}. Notably, the LLMs first convert the homophone into its pronunciation (pinyin) and then attempt to recover the target word, supporting our two-step hypothesis that LLMs perform phonological understanding via the sound layer (as discussed in \S~\ref{sssec:Analysis 1}).

\subsection{Similarity Comparison in Pinyin Form}
\label{aap:Similarity Comparison in pinyin}
\begin{figure}[!bt]
    \centering
    \includegraphics[width = \linewidth]{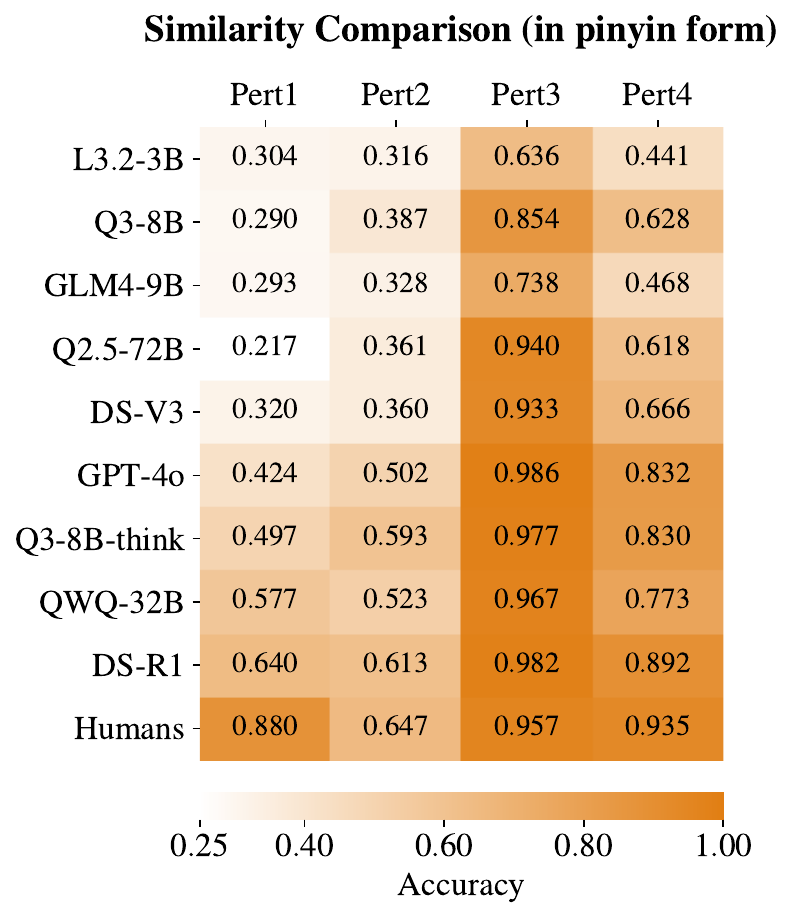}
    \caption{Results of the SC task in pinyin form. L, Q and DS denote \texttt{Llama}, \texttt{Qwen} and \texttt{DeepSeek}, respectively. For the human baseline, we still use results from the word-form setting.}
    \label{fig:D3_res_py}
\end{figure}

\begin{figure*}[th]
    \centering
    \includegraphics[width = \linewidth]{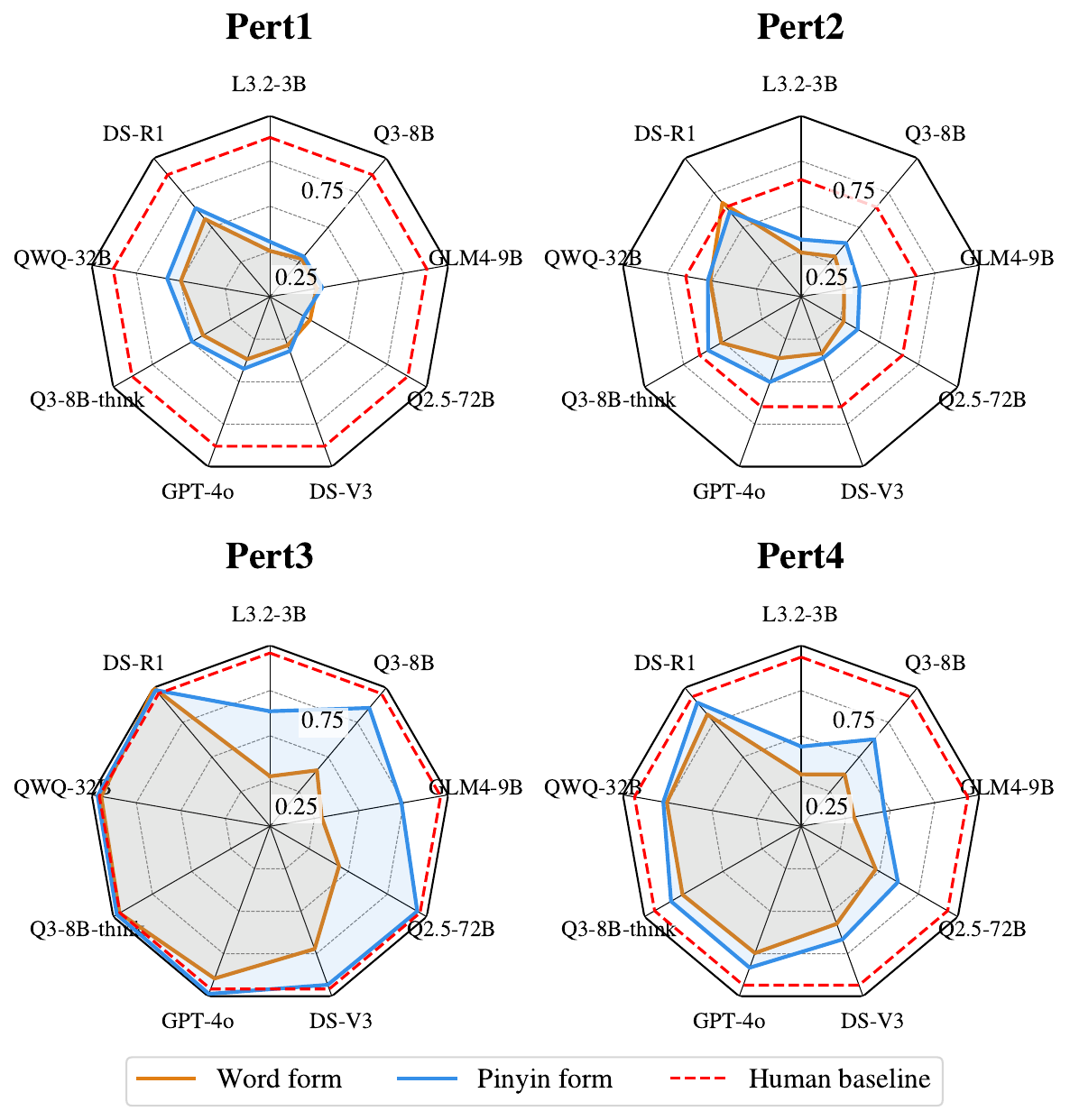}
    \caption{Comparison between performances under the pinyin-form and word-form settings. L, Q and DS denote \texttt{Llama}, \texttt{Qwen} and \texttt{DeepSeek}, respectively.}
    \label{fig:D3_radar}
\end{figure*}

We present the results of the SC task in pinyin form in Figure~\ref{fig:D3_res_py}, and a performance comparison between the two settings in Figure~\ref{fig:D3_radar}. We observe that the pinyin-form setting generally improves LLM performance\footnote{In contrast, human participants' performance drops, as they reported that the pinyin-form setting was difficult to read.}, consistent with our hypothesis that LLMs perform phonological understanding primarily via the sound layer. 

\subsection{Evaluation on Audio-Enabled LLMs}
\label{aap:evaluation on audio LLM}
We use \texttt{Qwen-TTS}\footnote{\url{https://www.alibabacloud.com/help/en/model-studio/qwen-tts}} to convert the questions into audio files. These audio files, combined with a short prompt (see Figure~\ref{fig:pmpt_speech}), are then evaluated by \texttt{Qwen-Omni-Turbo} and \texttt{Qwen3-Omni-Flash}. The speech templates and prompts are shown in Figure~\labelcref{fig:speech_speech2idiom,fig:pmpt_speech}. 

\begin{figure*}[!tb]
    \centering
    \includegraphics[width = \textwidth]{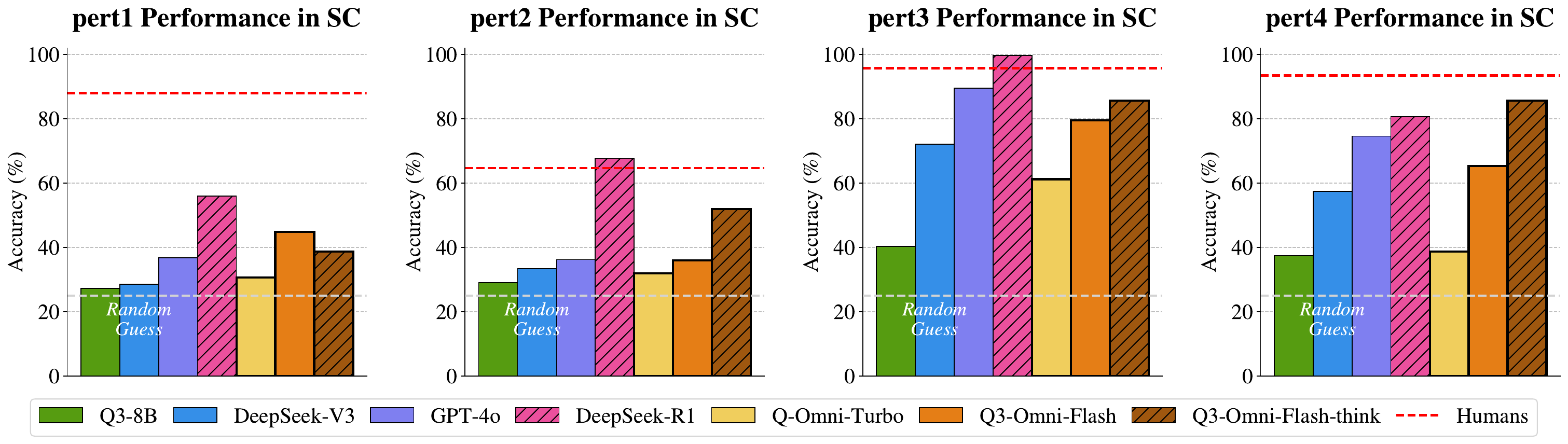}
    \caption{Similarity Comparison results for audio models across perturbation settings.}
    \label{fig:audio_pert_sc}
\end{figure*}

The results for each perturbation setting are shown in Figure~\ref{fig:audio_pert_sc}. Audio capability clearly improves performance; when combined with explicit reasoning, \texttt{Qwen3-Omni-Flash-think} even surpasses \texttt{DeepSeek-R1} in the $Pert_4$ setting. Interestingly, the relative strength of audio models differs from that of text-only LLMs: for example, LLMs perform better in the $Pert_3$ setting, likely because token-level similarity makes tone differences easy to capture in text form. In contrast, audio models excel in the $Pert_4$ setting, which is relatively easier to resolve acoustically, as also testified by human performance.

\section{Scaling and Mode Analysis of Qwen3 Models}
\label{app:Qwen3 Scaling}
To systematically analyze how phonological understanding emerges and scales with model size and reasoning mode, we conduct evaluations of our main tasks on the open-sourced \texttt{Qwen3} models, covering all eight available sizes/architectures (from 0.6B to 235B-A22B), in both non-thinking and thinking modes.

The results, summarized in Table~\ref{tab:qwen3_scaling} and Figure~\ref{fig:qwen_scaling}, demonstrate that:
\begin{itemize}
    \item \textbf{Performance scales with model size:} Across nearly all tasks, larger models consistently outperform smaller ones. For example, in the Similarity Comparison (SC) task, smaller models (0.6B and 1.7B) achieve near-random performance ($\sim0.25$) on $Pert_4$, but the capability clearly emerges as model size increases.
    \item \textbf{Benefit of long CoT (thinking mode):} The thinking mode significantly outperforms the non-thinking mode, demonstrating the clear benefit of explicit step-by-step reasoning via the sound layer.
\end{itemize}

Despite these improvements, even the largest model (235B-A22B) in thinking mode still shows room for improvement on certain challenging tasks, such as 2-rhyme Rhyming Sentence Generation (RSG). 

\begin{table*}[thb]
\setlength{\abovecaptionskip}{0pt}
\setlength{\belowcaptionskip}{-10pt}
\begin{center}
\resizebox{\linewidth}{!}{%
\small
\begin{tblr}{
colsep=6pt,
colspec={l*{7}{X[c]}X[1.8,c]*{8}{X[c]}}, 
row{1} = {font=\bfseries},
row{2} = {font=\bfseries},
column{1} = {font=\bfseries},
hline{1,Z} = {1pt,solid},
hline{2} = {2-9}{0.3pt, solid},
hline{2} = {10-17}{0.3pt, solid},
hline{3} = {0.5pt, solid},
hline{13} = {0.5pt, solid},
}
\SetCell[r=2]{c,m}\bf Task 
& \SetCell[c=8]{c,m}\bf Non-Thinking Mode &&&&&&& 
& \SetCell[c=8]{c,m}\bf Thinking Mode &&&&&&& \\ 
& \SetCell[c=1]{c,m} \bf 0.6B & \SetCell[c=1]{c,m} \bf 1.7B & \SetCell[c=1]{c,m} \bf 4B & \SetCell[c=1]{c,m} \bf 8B & \SetCell[c=1]{c,m} \bf 14B & \SetCell[c=1]{c,m} \bf 32B & \SetCell[c=1]{c,m} \bf 30B-A3B & \SetCell[c=1]{c,m} \bf 235B-A22B
& \SetCell[c=1]{c,m} \bf 0.6B & \SetCell[c=1]{c,m} \bf 1.7B & \SetCell[c=1]{c,m} \bf 4B & \SetCell[c=1]{c,m} \bf 8B & \SetCell[c=1]{c,m} \bf 14B & \SetCell[c=1]{c,m} \bf 32B & \SetCell[c=1]{c,m} \bf 30B-A3B & \SetCell[c=1]{c,m} \bf 235B-A22B \\

HR & 0.00 & 0.00 & 0.08 & 0.14 & 0.23 & 0.24 & 0.25 & 0.37 & 0.00 & 0.08 & 0.41 & 0.67 & 0.78 & 0.76 & 0.77 & 0.98 \\
CHR (DR) & 0.80 & 0.90 & 0.88 & 0.96 & 0.96 & 0.96 & 0.97 & 0.98 & 0.71 & 0.91 & 0.96 & 0.96 & 0.99 & 0.97 & 0.97 & 0.99 \\
CHR (RR) & 0.05 & 0.12 & 0.23 & 0.35 & 0.42 & 0.45 & 0.41 & 0.54 & 0.07 & 0.15 & 0.22 & 0.36 & 0.53 & 0.53 & 0.47 & 0.70 \\
RSG (1-Rhyme) & 0.07 & 0.10 & 0.16 & 0.39 & 0.49 & 0.60 & 0.40 & 0.56 & 0.13 & 0.14 & 0.48 & 0.65 & 0.77 & 0.74 & 0.71 & 0.78 \\
RSG (2-Rhyme) & 0.01 & 0.01 & 0.01 & 0.03 & 0.04 & 0.05 & 0.04 & 0.04 & 0.02 & 0.02 & 0.10 & 0.07 & 0.31 & 0.18 & 0.10 & 0.24 \\
SC ($Pert_1$) & 0.23 & 0.25 & 0.25 & 0.27 & 0.26 & 0.28 & 0.26 & 0.27 & 0.27 & 0.31 & 0.38 & 0.38 & 0.41 & 0.48 & 0.45 & 0.45 \\
SC ($Pert_2$) & 0.23 & 0.25 & 0.27 & 0.29 & 0.30 & 0.34 & 0.29 & 0.34 & 0.25 & 0.26 & 0.38 & 0.60 & 0.50 & 0.56 & 0.53 & 0.49 \\
SC ($Pert_3$) & 0.29 & 0.30 & 0.39 & 0.41 & 0.49 & 0.54 & 0.40 & 0.63 & 0.33 & 0.55 & 0.71 & 0.97 & 0.91 & 0.96 & 0.91 & 0.99 \\
SC ($Pert_4$) & 0.27 & 0.28 & 0.30 & 0.38 & 0.47 & 0.54 & 0.42 & 0.58 & 0.25 & 0.42 & 0.51 & 0.79 & 0.76 & 0.74 & 0.71 & 0.79 \\
SC (Ave.) & 0.25 & 0.27 & 0.30 & 0.34 & 0.38 & 0.42 & 0.34 & 0.45 & 0.27 & 0.38 & 0.49 & 0.68 & 0.65 & 0.68 & 0.65 & 0.68 \\
Overall & 0.09 & 0.12 & 0.19 & 0.31 & 0.38 & 0.43 & 0.35 & 0.48 & 0.12 & 0.19 & 0.40 & 0.59 & 0.68 & 0.68 & 0.65 & 0.78 \\
\end{tblr}
}
\end{center}
\caption{Performance scaling of \texttt{Qwen3} models across different sizes and modes. The overall score is computed as the average of HR, CHR~(RR), RSG~(1-Rhyme), RSG~(2-Rhyme), and SC~(Ave.).}
\label{tab:qwen3_scaling}
\end{table*}

\begin{figure*}[t]
\setlength{\abovecaptionskip}{5pt}
\setlength{\belowcaptionskip}{-12pt}
    \centering
    \includegraphics[width=0.8\linewidth]{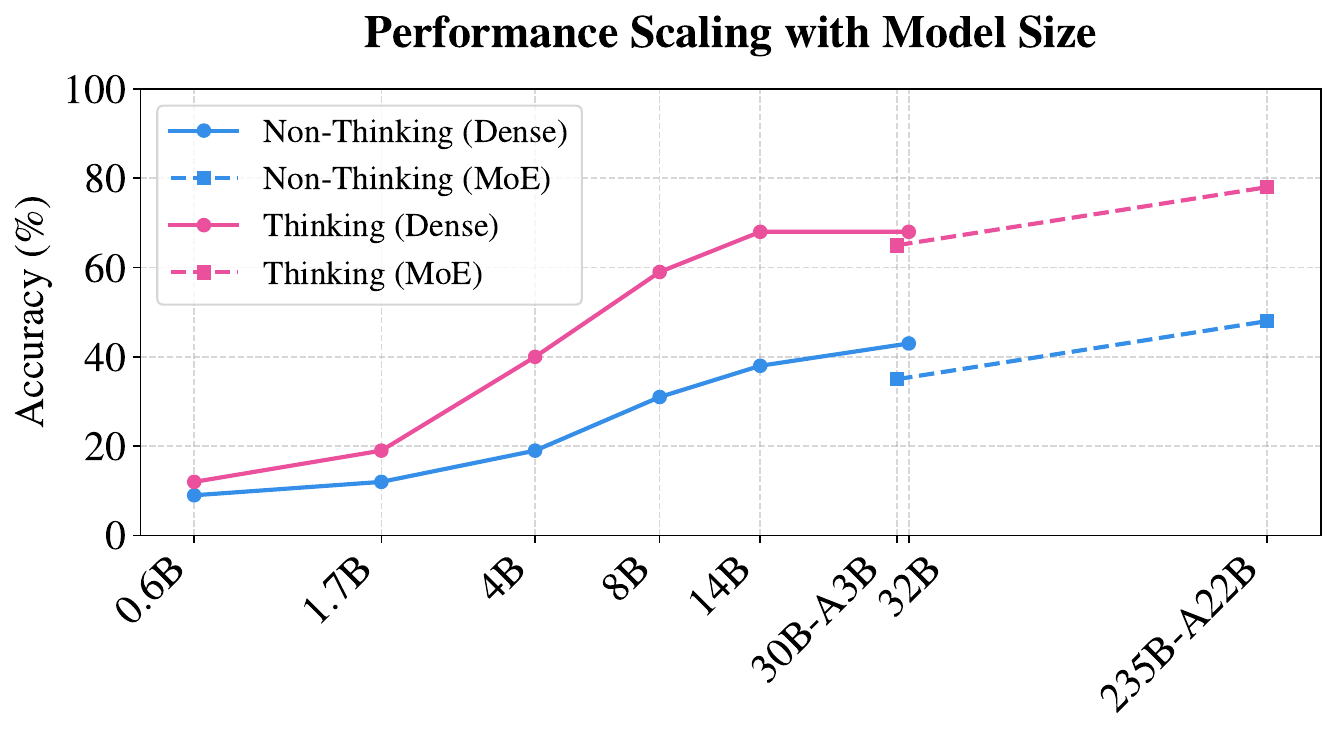}
    \caption{Overall performance scaling of \texttt{Qwen3} models across different sizes (0.6B to 235B-A22B) and modes (Non-Thinking vs. Thinking).}
    \label{fig:qwen_scaling}
\end{figure*}

\section{Prompt Templates and Human Instructions}
\label{app:Prompts and Instructions}
The prompt templates used in model evaluations are as follows:
\begin{itemize}
    \item \textbf{Verification Experiment}: See Figure~\ref{fig:pmpt_verify_char2py}.
    \item \textbf{Homophony-Based Understanding}: See Figure~\labelcref{fig:pmpt_hu_hc_homo2idiom,fig:pmpt_hu_chr_pun}.
    \item \textbf{RQ~1}: See Figure~\labelcref{fig:pmpt_verify_py2char,fig:pmpt_verify_char2char,fig:pmpt_hu_hc_homo2py,fig:pmpt_hu_hc_py2idiom,fig:pmpt_hu_chr_phrase2py,fig:pmpt_hu_chr_py2phrase}.
    \item \textbf{Rhyme-Based Understanding and RQ~2:} See Figure~\labelcref{fig:pmpt_ru_r1,fig:pmpt_ru_r2}.
    \item \textbf{Phonetic-Similarity-Based Understanding and RQ~3:} See Figure~\labelcref{fig:pmpt_psu_sc_word4word,fig:pmpt_psu_sc_py4py}.
\end{itemize}

The speech and prompt templates used in the audio-model experiment (\S~\ref{sssec:Extended Experiment}) are in Figure~\labelcref{fig:speech_speech2idiom,fig:pmpt_speech}.

\begin{figure*}
    \centering
    \includegraphics[width = \linewidth]{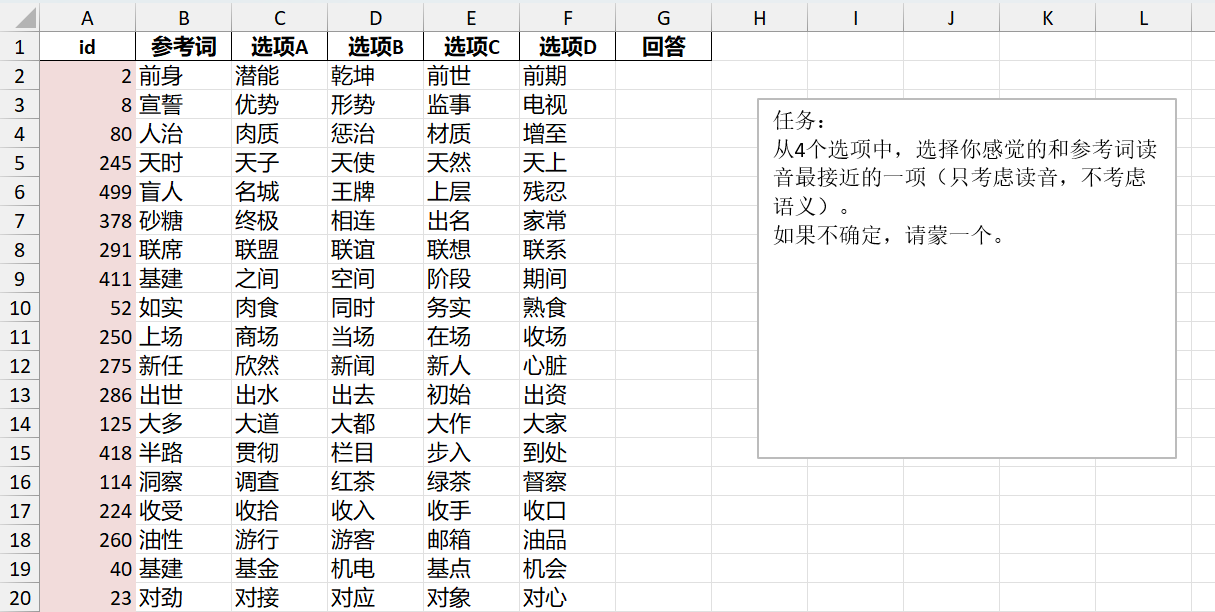}
    \caption{Screenshot of the Similarity Comparison \texttt{.xlsx} file provided to human participants.}
    \label{fig:human_interface}
\end{figure*}

The human instructions are shown in Figures~\labelcref{fig:inst_verify_char2py,fig:inst_hu_hc_homo2idiom,fig:inst_hu_chr_pun,fig:inst_psu_sc_word4word,fig:inst_psu_sc_py4py}, and a screenshot of an \texttt{.xlsx} file provided to participants is shown in Figure~\ref{fig:human_interface}.

\begin{figure*}[ptb]
\centering
\begin{tcolorbox}

\textbf{Case Study on Homophone Recall \textcolor{true}{(Correct)}}
\medskip\hrule\medskip

\begin{CJK}{UTF8}{gbsn}
\textbf{Prompt:}\\
你是一位精通汉语的大师。请根据下面给出的一个“短语”，猜出和它在读音上最相近的四字成语（不要考虑语义），以json格式输出答案。示例如下：
\begin{verbatim}
<content>孰语盆笔</content>
输出：{
    "answer": "无与伦比"
}
<content>律师捕霜</content>
输出：{
    "answer": "屡试不爽"
}
<content>王杨捕劳</content>
输出：{
    "answer": "亡羊补牢"
}
\end{verbatim}
本题目给出的短语如下：\\
\verb|<content>|\texttt{\textcolor{brown7}{呻鳞骑竟}}\verb|</content>|

\vspace{0.5cm}
\textbf{Thought:}
\begin{verbatim}
首先，用户提供了短语："呻鳞骑竟"。我需要根据它的读音猜出最相近的四字成语，只考虑读音，不考虑语义。输出必须是JSON格式，如示例所示。
\end{verbatim}
\textcolor{gray6}{\emph{(... analysis on examples omitted)}}
\begin{verbatim}
现在，我的短语是 "呻鳞骑竟"。我需要找出它在读音上最接近的四字成语。
先分解 "呻鳞骑竟" 的发音：
- "呻"：拼音是 "shēn"（shēn）。
- "鳞"：拼音是 "lín"（lín）。
- "骑"：拼音是 "qí"（qí）。
- "竟"：拼音是 "jìng"（jìng）。
\end{verbatim}
所以，整个短语的发音是 \textcolor{true}{"shēn lín qí jìng"}。
\begin{verbatim}
我需要找到一个四字成语，其发音最接近 "shēn lín qí jìng"。
回想常见的四字成语：
\end{verbatim}
- 可能成语有 \textcolor{true}{"身临其境"}？"身临其境" 发音是 "shēn lín qí jìng"。\\
\textcolor{gray6}{\emph{(... verification and more guesses omitted)}}
\begin{verbatim}
现在，输出格式必须是JSON：
{
    "answer": "身临其境"
}
在示例中，是完整的JSON对象。
最后，确保只输出JSON，没有其他文本。
\end{verbatim}

\vspace{0.5cm}
\textbf{Answer:}
\textcolor{true}{"身临其境"}
\end{CJK}
\end{tcolorbox}
\caption{A correct reasoning trace for the Homophone Recall task from \texttt{DeepSeek-R1}.}
\label{fig:output_hr_ds}
\end{figure*}
\begin{figure*}[ptb]
\centering
\begin{tcolorbox}

\textbf{Case Study on Homophone Recall \textcolor{false}{(Incorrect)}}
\medskip\hrule\medskip

\begin{CJK}{UTF8}{gbsn}
\textbf{Prompt:}\\
你是一位精通汉语的大师。请根据下面给出的一个“短语”，猜出和它在读音上最相近的四字成语（不要考虑语义），以json格式输出答案。示例如下：
\begin{verbatim}
<content>孰语盆笔</content>
输出：{
    "answer": "无与伦比"
}
<content>律师捕霜</content>
输出：{
    "answer": "屡试不爽"
}
<content>王杨捕劳</content>
输出：{
    "answer": "亡羊补牢"
}
\end{verbatim}
本题目给出的短语如下：\\
\verb|<content>|\texttt{\textcolor{brown7}{案怖臼颁}}\verb|</content>|

\vspace{0.5cm}
\textbf{Thought:}
\begin{verbatim}
好的，我现在需要解决用户的问题，根据给出的拼音“案怖臼颁”来猜测一个读音相近的四字成语。首先，我需要将每个拼音拆解，然后寻找对应的汉字，再组合成成语。
首先，分析每个拼音对应的可能汉字：
1. “案”拼音是an，可能的汉字有案、安、暗等。
2. “怖”拼音是bu，可能的汉字有怖、布、步等。
3. “臼”拼音是jiu，可能的汉字有臼、旧、就等。
4. “颁”拼音是ban，可能的汉字有颁、班、搬等。
接下来，我需要将这些可能的汉字组合成四字成语，并且每个字的拼音要与原短语中的拼音一致。例如，第一个字是an，第二个是bu，第三个是jiu，第四个是ban。那么可能的组合是：安布旧班？但这样组合起来似乎没有意义，可能需要调整每个字的可能选项。
或者，可能每个拼音对应的字不是直接对应，而是每个字的拼音与原短语中的拼音相同，但字形不同。例如，“案”可能对应“安”，“怖”对应“布”，“臼”对应“旧”，“颁”对应“班”，这样组合成“安布旧班”，但这样的成语不存在。
可能我需要重新考虑每个拼音对应的字。例如，“案”可能对应“按”，但拼音是an，所以可能没问题。或者“怖”可能对应“捕”（bu），但拼音是bu，所以可能。然后“臼”可能对应“就”（jiu），而“颁”对应“班”（ban）。这样组合起来是“按捕就班”？这似乎也不对。
\end{verbatim}
\verb|或者，可能每个字的拼音完全对应，但需要找到四字成语。例如，原短语是an、bu、jiu、ban。有没有成语的拼音是|\textcolor{true}{an bu jiu ban}？比如\textcolor{false}{“安步就班”}？安步就班的拼音是an bu jiu ban，正好对应。所以答案可能是“安步就班”。
\begin{verbatim}
检查一下这个成语是否存在。“安步就班”确实是一个成语，意思是安步当车，就班取舍，形容安详从容，不慌不忙。所以这应该符合要求。因此，正确的答案应该是“安步就班”。
\end{verbatim}

\vspace{0.5cm}
\textbf{Answer:}
\textcolor{false}{"安步就班"}
\end{CJK}
\end{tcolorbox}
\caption{An incorrect reasoning trace for the Homophone Recall task from \texttt{Qwen3-8B-think}. The correct answer is ``\begin{CJK}{UTF8}{gbsn}按部就班\end{CJK}''.}
\label{fig:output_hr_q}
\end{figure*}
\begin{figure*}[ptb]
\centering
\begin{tcolorbox}[
    fontupper=\small
]

\textbf{Case Study on Similarity Comparison \textcolor{true}{(Correct)}}
\medskip\hrule\medskip

\begin{CJK}{UTF8}{gbsn}
\textbf{Prompt:}\\
你是一位精通汉语的大师。请从四个选项中，选出在读音上和给出的词语最接近的一项（不要考虑语义），以json格式输出答案。示例如下：\\
\textcolor{gray6}{\emph{(... examples omitted)}}\\
本题目给出的词语和选项如下：\\
\verb|<content>|\texttt{\textcolor{brown7}{摊位}}\verb|</content>|\\
\verb|<choices>A. 环卫 B. 调味 C. 因为 D. 单位</choices>|

\vspace{0.5cm}
\textbf{Thought:}
\begin{verbatim}
首先，用户要求我选出在读音上和给出的词语“摊位”最接近的一项。不要考虑语义，只关注发音。输出必须是JSON格式，如{"answer": "选项"}。
给出的词语是“摊位”，拼音是“tān wèi”。
现在，分析每个选项的拼音：
- A. 环卫：拼音是“huán wèi”。
- B. 调味：拼音是“tiáo wèi”。
- C. 因为：拼音是“yīn wèi”。
- D. 单位：拼音是“dān wèi”。
目标词语“摊位”的拼音是“tān wèi”。我需要找出哪个选项的发音最接近“tān wèi”。
比较每个选项：
- A. 环卫：“huán wèi” – 声母“h” vs “t”，韵母“uán” vs “ān”，完全不同。韵尾“wèi”相同，但开头差异大。
- B. 调味：“tiáo wèi” – 声母“t”相同，但韵母“iáo” vs “ān”。韵尾“wèi”相同，但“iáo”和“ān”发音不同。
- C. 因为：“yīn wèi” – 声母“y” vs “t”，韵母“īn” vs “ān”。韵尾“wèi”相同，但开头差异大。
- D. 单位：“dān wèi” – 声母“d” vs “t”，韵母“ān”相同，韵尾“wèi”相同。“d”和“t”都是舌尖音，但“d”是浊音，“t”是清音。在汉语中，“d”和“t”发音相似，都是齿龈塞音，一个浊一个清，但音位接近。韵母“ān”完全一样。
重点比较“摊位”(tān wèi) 和 “单位”(dān wèi)：
- 声母：t vs d – 都是齿龈塞音，t是送气清音，d是不送气浊音。在普通话中，它们发音相似，容易混淆，尤其在某些口音中。
- 韵母：ān 完全相同。
- 韵尾：wèi 完全相同。
所以，“摊位”和“单位”在发音上非常接近，主要区别在声母t和d。
现在看其他选项：
- B. 调味 (tiáo wèi)：声母t相同，但韵母iáo vs ān，差异更大。
- A. 环卫 (huán wèi)：声母和韵母都不同。
- C. 因为 (yīn wèi)：声母和韵母不同。
因此，D. 单位 是最接近的。
回顾之前的例子：
- 在“增添”的例子中，选了“春天”，因为“增添”(zēng tiān) 和 “春天”(chūn tiān) 都有“tiān”，尽管声母不同。
- 在“渐渐”的例子中，选了“见面”，因为“渐渐”(jiàn jiàn) 和 “见面”(jiàn miàn) 共享“jiàn”。
类似地，这里“摊位”和“单位”共享“wèi”，并且“ān”相同，声母t和d相似。
在汉语中，“摊位”读作 tān wèi，“单位”读作 dān wèi。发音上，t和d在音位上对立，但音值接近，尤其在不强调时。
我认为D是最佳选择。
确认输出格式：必须是JSON，如{"answer": "D"}。
所以，我的输出应该是{"answer": "D"}。
\end{verbatim}
\vspace{0.5cm}
\textbf{Answer:}
\textcolor{true}{"D"}
\end{CJK}
\end{tcolorbox}
\caption{A correct reasoning trace for the Similarity Comparison task ($Pert_1$ setting) from \texttt{DeepSeek-R1}.}
\label{fig:output_sc_ds}
\end{figure*}
\begin{figure*}[ptb]
\centering
\begin{tcolorbox}[
]

\textbf{Case Study on Similarity Comparison \textcolor{false}{(Incorrect)}}
\medskip\hrule\medskip

\begin{CJK}{UTF8}{gbsn}
\textbf{Prompt:}\\
你是一位精通汉语的大师。请从四个选项中，选出在读音上和给出的词语最接近的一项（不要考虑语义），以json格式输出答案。示例如下：\\
\textcolor{gray6}{\emph{(... examples omitted)}}\\
本题目给出的词语和选项如下：\\
\verb|<content>|\texttt{\textcolor{brown7}{今生}}\verb|</content>|\\
\verb|<choices>A. 晋升 B. 发生 C. 民生 D. 寄生</choices>|

\vspace{0.5cm}
\textbf{Thought:}
\begin{verbatim}
好的，我现在需要解决这个题目，找出与“今生”在读音上最接近的选项。首先，我要仔细分析每个选项的拼音，然后与“今生”的拼音进行比较。
首先，“今生”的拼音是“jīn shēng”。接下来，我要逐个分析选项：
\end{verbatim}
\verb|选项A是“晋升”，拼音是“jìn shēn”。这里要注意的是，“晋升”的“晋”发音是jìn，而“今生”的“今”是jīn，这|\textcolor{false}{两个韵母不同}\verb|，声调也不同（“今”是第一声，“晋”是第四声）。所以A选项的发音和“今生”不太接近。|
\begin{verbatim}
选项B是“发生”，拼音是“fā shēng”。“发生”的“发”是fā，而“今生”的“今”是jīn，声母不同，韵母部分“生”是shēng，而“今生”的“生”也是shēng，但前面的声母不同，所以整体发音差异较大。
选项C是“民生”，拼音是“mín shēng”。“民生”的“民”是mín，而“今生”的“今”是jīn，声母不同，但韵母都是“in”结尾，不过“今生”的“生”是shēng，而“民生”的“生”是shēng，所以韵母部分相同，但声母不同。不过可能需要考虑整体发音的接近程度。
选项D是“寄生”，拼音是“jì shēng”。“寄生”的“寄”是jì，而“今生”的“今”是jīn，声母都是j，但韵母不同，“今”是in，“寄”是i，不过声调方面，“今”是第一声，“寄”是第四声。不过“生”部分都是shēng，所以后半部分相同。
现在比较各个选项与“今生”的发音接近程度。首先看声母部分，“今生”的声母是j（jīn）和sh（shēng）。选项C的“民生”声母是m和sh，选项D的“寄生”是j和sh。所以D选项的声母部分与“今生”的前一个字“今”（j）相同，而“生”部分相同，所以整体发音更接近。而选项C的“民”是m，与j不同。因此，D选项的“寄生”可能更接近。
不过需要再仔细检查。例如，“今生”是jīn shēng，而“寄生”是jì shēng。前一个字的韵母是in和i，声调不同，但声母相同。后一个字完全相同。因此，可能D选项的发音更接近，因为前一个字的声母相同，后一个字完全相同，尽管韵母和声调不同，但整体可能更接近。而选项C的“民生”是mín shēng，前一个字声母不同，但韵母相同，但声母差异较大。因此，D选项更接近。
\end{verbatim}
\vspace{0.5cm}
\textbf{Answer:}
\textcolor{false}{"D"}
\end{CJK}
\end{tcolorbox}
\caption{An incorrect reasoning trace for the Similarity Comparison task ($Pert_3$ setting) from \texttt{Qwen3-8B-think}. The correct answer is ``A''.}
\label{fig:output_sc_q}
\end{figure*}

\begin{figure*}[ptb]
\centering
\begin{tcolorbox}

\textbf{Prompt Template}
\medskip\hrule\medskip

\begin{CJK}{UTF8}{gbsn}
\textbf{Chinese:}\\
你是一位精通汉语的大师。请根据下面给出的汉字，写出它所对应的标准普通话拼音，以json格式输出答案。示例如下：
\begin{verbatim}
<content>张</content>
输出：{
    "answer": "zhāng"
}
<content>我</content>
输出：{
    "answer": "wǒ"
}
<content>聂</content>
输出：{
    "answer": "niè"
}
\end{verbatim}
本题目给出的汉字如下：\\
\verb|<content>|\texttt{\textcolor{brown7}{\{content\}}}\verb|</content>|

\vspace{0.5cm}
\textbf{English Translation:}\\
You are a master of Chinese. Please provide the standard Mandarin pinyin for the Chinese characters given below, and output the answer in JSON format. Example:
\begin{verbatim}
<content>张</content>
Output: {
    "answer": "zhāng"
}
<content>我</content>
Output: {
    "answer": "wǒ"
}
<content>聂</content>
Output: {
    "answer": "niè"
}
\end{verbatim}
The Chinese character given in this problem is as follows:\\
\verb|<content>|\texttt{\textcolor{brown7}{\{content\}}}\verb|</content>|
\end{CJK}
\end{tcolorbox}
\caption{Prompt template for the character-to-pinyin verification experiment (\S~\ref{ssec:Verification experiment}). \texttt{\textcolor{brown7}{\{content\}}} is replaced with the actual character.}
\label{fig:pmpt_verify_char2py}
\end{figure*}

\begin{figure*}[ptb]
\centering
\begin{tcolorbox}

\textbf{Prompt Template}
\medskip\hrule\medskip

\begin{CJK}{UTF8}{gbsn}
\textbf{Chinese:}\\
你是一位精通汉语的大师。请根据下面给出的一个“短语”，猜出和它在读音上最相近的四字成语（不要考虑语义），以json格式输出答案。示例如下：
\begin{verbatim}
<content>孰语盆笔</content>
输出：{
    "answer": "无与伦比"
}
<content>律师捕霜</content>
输出：{
    "answer": "屡试不爽"
}
<content>王杨捕劳</content>
输出：{
    "answer": "亡羊补牢"
}
\end{verbatim}
本题目给出的短语如下：\\
\verb|<content>|\texttt{\textcolor{brown7}{\{content\}}}\verb|</content>|

\vspace{0.5cm}
\textbf{English Translation:}\\
You are a master of Chinese. Based on the given “phrase,” guess a four-character idiom whose pronunciation is closest to it (ignore meaning), and output the answer in JSON format. Example:
\begin{verbatim}
<content>孰语盆笔</content>
输出：{
    "answer": "无与伦比"
}
<content>律师捕霜</content>
输出：{
    "answer": "屡试不爽"
}
<content>王杨捕劳</content>
输出：{
    "answer": "亡羊补牢"
}
\end{verbatim}
The phrase given in this problem is as follows:\\
\verb|<content>|\texttt{\textcolor{brown7}{\{content\}}}\verb|</content>|
\end{CJK}
\end{tcolorbox}
\caption{Prompt template for the Homophone Recall task (\S~\ref{sssec:Homophone Recall}). \texttt{\textcolor{brown7}{\{content\}}} is replaced with the actual homophonic ``phrase''.}
\label{fig:pmpt_hu_hc_homo2idiom}
\end{figure*}
\begin{figure*}[ptb]
\centering
\begin{tcolorbox}

\textbf{Prompt Template}
\medskip\hrule\medskip

\begin{CJK}{UTF8}{gbsn}
\textbf{Chinese:}\\
你是一位精通汉语的大师。请从下面给出的包含谐音梗的句子中，抽取出关键的谐音部分，并写出它所对应的常用语（包括词语、成语），以json格式输出答案。示例如下：
\begin{verbatim}
<sentence>红色的圆形印章，简单来说，就是，朱圆章。</sentence>
输出：{
    "pun word": "朱圆章",
    "alternative word": "朱元璋"
}
<sentence>龙下凡，我接他，他说需要用雨水才能接他，于是我盛雨接龙。</sentence>
输出：{
    "pun word": "盛雨接龙",
    "alternative word": "成语接龙"
}
<sentence>据理力争的女律师被评为当代据理夫人。</sentence>
输出：{
    "pun word": "据理夫人",
    "alternative word": "居里夫人"
}
\end{verbatim}
本题目给出的句子如下：\\
\verb|<sentence>|\texttt{\textcolor{green7}{\{sentence\}}}\verb|</sentence>|

\vspace{0.5cm}
\textbf{English Translation:}\\
You are a master of Chinese. Please extract the key homophonic part from the given sentence that contains a pun, and provide the corresponding common word or idiom. Output the answer in JSON format. Example:
\begin{verbatim}
<sentence>红色的圆形印章，简单来说，就是，朱圆章。</sentence>
输出：{
    "pun word": "朱圆章",
    "alternative word": "朱元璋"
}
<sentence>龙下凡，我接他，他说需要用雨水才能接他，于是我盛雨接龙。</sentence>
输出：{
    "pun word": "盛雨接龙",
    "alternative word": "成语接龙"
}
<sentence>据理力争的女律师被评为当代据理夫人。</sentence>
输出：{
    "pun word": "据理夫人",
    "alternative word": "居里夫人"
}
\end{verbatim}
The sentence given in this problem is as follows:\\
\verb|<sentence>|\texttt{\textcolor{green7}{\{sentence\}}}\verb|</sentence>|
\end{CJK}
\end{tcolorbox}
\caption{Prompt template for the Contextual Homophone Recognition task (\S~\ref{sssec:Contextual Homophone Recognition}). \texttt{\textcolor{green7}{\{sentence\}}} is replaced with the actual pun sentence.}
\label{fig:pmpt_hu_chr_pun}
\end{figure*}

\begin{figure*}[ptb]
\centering
\begin{tcolorbox}

\textbf{Prompt Template}
\medskip\hrule\medskip

\begin{CJK}{UTF8}{gbsn}
\textbf{Chinese:}\\
你是一位精通汉语的大师。请根据下面给出的拼音，写出5个发音和该拼音完全相同的汉字，以json格式输出答案。示例如下：
\begin{verbatim}
<pinyin>chén</pinyin>
输出：{
    "answer": ["晨", "辰", "沉", "尘", "陈"]
}
<pinyin>jīng</pinyin>
输出：{
    "answer": ["茎", "鲸", "荆", "晶", "兢"]
}
<pinyin>yì</pinyin>
输出：{
    "answer": ["艺", "抑", "译", "忆", "易"]
}
\end{verbatim}
本题目给出的拼音如下：\\
\verb|<pinyin>|\texttt{\textcolor{brown7}{\{content\}}}\verb|</pinyin>|

\vspace{0.5cm}
\textbf{English Translation:}\\
You are a master of Chinese. Please provide 5 Chinese characters that share exactly the same pronunciation as the given pinyin, and output the answer in JSON format. Example:
\begin{verbatim}
<pinyin>chén</pinyin>
Output：{
    "answer": ["晨", "辰", "沉", "尘", "陈"]
}
<pinyin>jīng</pinyin>
Output：{
    "answer": ["茎", "鲸", "荆", "晶", "兢"]
}
<pinyin>yì</pinyin>
Output：{
    "answer": ["艺", "抑", "译", "忆", "易"]
}
\end{verbatim}
The pinyin given in this problem is as follows:\\
\verb|<pinyin>|\texttt{\textcolor{brown7}{\{content\}}}\verb|</pinyin>|
\end{CJK}
\end{tcolorbox}
\caption{Prompt template for the pinyin2char test in RQ~1 (\S~\ref{app:RQ1}, for the Character-Level Homophone Recall task). \texttt{\textcolor{brown7}{\{content\}}} is replaced with the actual pinyin.}
\label{fig:pmpt_verify_py2char}
\end{figure*}
\begin{figure*}[ptb]
\centering
\begin{tcolorbox}

\textbf{Prompt Template}
\medskip\hrule\medskip

\begin{CJK}{UTF8}{gbsn}
\textbf{Chinese:}\\
你是一位精通汉语的大师。请根据下面给出的汉字，写出5个发音和它完全相同的汉字，以json格式输出答案。示例如下：
\begin{verbatim}
<content>臣</content>
输出：{
    "answer": ["晨", "辰", "沉", "尘", "陈"]
}
<content>京</content>
输出：{
    "answer": ["茎", "鲸", "荆", "晶", "兢"]
}
<content>义</content>
输出：{
    "answer": ["艺", "抑", "译", "忆", "易"]
}
\end{verbatim}
本题目给出的汉字如下：\\
\verb|<content>|\texttt{\textcolor{brown7}{\{content\}}}\verb|</content>|

\vspace{0.5cm}
\textbf{English Translation:}\\
You are a master of Chinese. Please provide 5 Chinese characters that share exactly the same pronunciation as the given character, and output the answer in JSON format. Example:
\begin{verbatim}
<content>臣</content>
Output：{
    "answer": ["晨", "辰", "沉", "尘", "陈"]
}
<content>京</content>
Output：{
    "answer": ["茎", "鲸", "荆", "晶", "兢"]
}
<content>义</content>
Output：{
    "answer": ["艺", "抑", "译", "忆", "易"]
}
\end{verbatim}
The character given in this problem is as follows:\\
\verb|<content>|\texttt{\textcolor{brown7}{\{content\}}}\verb|</content>|
\end{CJK}
\end{tcolorbox}
\caption{Prompt template for the char2pinyin test in RQ~1 (\S~\ref{app:RQ1}, for the Character-Level Homophone Recall task). \texttt{\textcolor{brown7}{\{content\}}} is replaced with the actual character.}
\label{fig:pmpt_verify_char2char}
\end{figure*}
\begin{figure*}[ptb]
\centering
\begin{tcolorbox}

\textbf{Prompt Template}
\medskip\hrule\medskip

\begin{CJK}{UTF8}{gbsn}
\textbf{Chinese:}\\
你是一位精通汉语的大师。请根据下面给出的一个“短语”，写出它所对应的标准普通话拼音(各汉字对应的拼音之间以空格隔开），以json格式输出答案。示例如下：
\begin{verbatim}
<content>吴雨仑笔</content>
输出：{
    "answer": "wú yǔ lún bǐ"
}
<content>履弑布爽</content>
输出：{
    "answer": "lǚ shì bù shuǎng"
}
<content>哆咄易鳝</content>
输出：{
    "answer": "duō duō yì shàn"
}
\end{verbatim}
本题目给出的短语如下：\\
\verb|<content>|\texttt{\textcolor{brown7}{\{content\}}}\verb|</content>|

\vspace{0.5cm}
\textbf{English Translation:}\\
You are a master of Chinese. Please provide the standard Mandarin pinyin for the given “phrase,” with each Chinese character’s pinyin separated by a space, and output the answer in JSON format. Example:
\begin{verbatim}
<content>吴雨仑笔</content>
Output：{
    "answer": "wú yǔ lún bǐ"
}
<content>履弑布爽</content>
Output：{
    "answer": "lǚ shì bù shuǎng"
}
<content>哆咄易鳝</content>
Output：{
    "answer": "duō duō yì shàn"
}
\end{verbatim}
The phrase given in this problem is as follows:\\
\verb|<content>|\texttt{\textcolor{brown7}{\{content\}}}\verb|</content>|
\end{CJK}
\end{tcolorbox}
\caption{Prompt template for the word2pinyin test in RQ~1 (\S~\ref{app:RQ1}, for the Homophone Recall task). \texttt{\textcolor{brown7}{\{content\}}} is replaced with the actual homophonic ``phrase''.}
\label{fig:pmpt_hu_hc_homo2py}
\end{figure*}
\begin{figure*}[ptb]
\centering
\begin{tcolorbox}

\textbf{Prompt Template}
\medskip\hrule\medskip

\begin{CJK}{UTF8}{gbsn}
\textbf{Chinese:}\\
\begin{verbatim}
你是一位精通汉语的大师。请根据下面给出的拼音，写出它所对应的成语，以json格式输出答案。示例如下：
<pinyin>wú yǔ lún bǐ</pinyin>
输出：{
    "answer": "无与伦比"
}
<pinyin>lǚ shì bù shuǎng</pinyin>
输出：{
    "answer": "屡试不爽"
}
<pinyin>duō duō yì shàn</pinyin>
输出：{
    "answer": "多多益善"
}
\end{verbatim}
本题目给出的拼音如下：\\
\verb|<pinyin>|\texttt{\textcolor{brown7}{\{pinyin\}}}\verb|</pinyin>|

\vspace{0.5cm}
\textbf{English Translation:}\\
You are a master of Chinese. Please guess a four-character idiom whose pronunciation is closest to the given pinyin, and output the answer in JSON format. Example:
\begin{verbatim}
<pinyin>wú yǔ lún bǐ</pinyin>
Output：{
    "answer": "无与伦比"
}
<pinyin>lǚ shì bù shuǎng</pinyin>
Output：{
    "answer": "屡试不爽"
}
<pinyin>duō duō yì shàn</pinyin>
Output：{
    "answer": "多多益善"
}
\end{verbatim}
The pinyin given in this problem is as follows:\\
\verb|<pinyin>|\texttt{\textcolor{brown7}{\{pinyin\}}}\verb|</pinyin>|
\end{CJK}
\end{tcolorbox}
\caption{Prompt template for the pinyin2word test in RQ~1 (\S~\ref{app:RQ1}, for the Homophone Recall task). \texttt{\textcolor{brown7}{\{pinyin\}}} is replaced with the actual pinyin.}
\label{fig:pmpt_hu_hc_py2idiom}
\end{figure*}
\begin{figure*}[ptb]
\centering
\begin{tcolorbox}

\textbf{Prompt Template}
\medskip\hrule\medskip

\begin{CJK}{UTF8}{gbsn}
\textbf{Chinese:}\\
你是一位精通汉语的大师。请根据下面给出的句子以及从句子中提取出来的短语，写出短语在句子中的语境下所对应的标准普通话拼音(各汉字对应的拼音之间以空格隔开），以json格式输出答案。示例如下：
\begin{verbatim}
<sentence>大户人家的小王犯了两项家规，父亲决定对小王数罪并罚，也就是家法结合律。</sentence>
<content>家法结合律</content>
输出：{
    "answer": "jiā fǎ jié hé lǜ"
}
<sentence>有人踩了笛卡尔一脚，笛卡尔：没事儿，因为我是恕学家。</sentence>
<content>恕学家</content>
输出：{
    "answer": "shù xué jiā"
}
<sentence>食材越多，越容易作出更多的菜，或许这就是多多益膳。</sentence>
<content>多多益膳</content>
输出：{
    "answer": "duō duō yì shàn"
}
\end{verbatim}
本题目给出的句子和短语如下：\\
\verb|<sentence>|\texttt{\textcolor{green7}{\{sentence\}}}\verb|</sentence>|\\
\verb|<content>|\texttt{\textcolor{brown7}{\{content\}}}\verb|</content>|

\vspace{0.5cm}
\textbf{English Translation:}\\
You are a master of Chinese. Please provide the standard Mandarin pinyin for the given phrase extracted from the sentence, in the context of that sentence, with each Chinese character’s pinyin separated by a space, and output the answer in JSON format. Example:
\begin{verbatim}
<sentence>大户人家的小王犯了两项家规，父亲决定对小王数罪并罚，也就是家法结合律。</sentence>
<content>家法结合律</content>
输出：{
    "answer": "jiā fǎ jié hé lǜ"
}
<sentence>有人踩了笛卡尔一脚，笛卡尔：没事儿，因为我是恕学家。</sentence>
<content>恕学家</content>
输出：{
    "answer": "shù xué jiā"
}
<sentence>食材越多，越容易作出更多的菜，或许这就是多多益膳。</sentence>
<content>多多益膳</content>
输出：{
    "answer": "duō duō yì shàn"
}
\end{verbatim}
The sentence and phrase given in this problem are as follows:\\
\verb|<sentence>|\texttt{\textcolor{green7}{\{sentence\}}}\verb|</sentence>|\\
\verb|<content>|\texttt{\textcolor{brown7}{\{content\}}}\verb|</content>|
\end{CJK}
\end{tcolorbox}
\caption{Prompt template for the word2pinyin test in RQ~1 (\S~\ref{app:RQ1}, for the Contextual Homophone Recognition task). \texttt{\textcolor{green7}{\{sentence\}}} is replaced with the actual context sentence, and \texttt{\textcolor{brown7}{\{content\}}} with the actual word.}
\label{fig:pmpt_hu_chr_phrase2py}
\end{figure*}
\begin{figure*}[ptb]
\centering
\begin{tcolorbox}

\textbf{Prompt Template}
\medskip\hrule\medskip

\begin{CJK}{UTF8}{gbsn}
\textbf{Chinese:}\\
你是一位精通汉语的大师。请根据下面给出的拼音，写出它所对应的最常用的短语（包括词语、成语），以json格式输出答案。示例如下：
\begin{verbatim}
<pinyin>jiā fǎ jié hé lǜ</pinyin>
输出：{
    "answer": "加法结合律"
}
<pinyin>shù xué jiā</pinyin>
输出：{
    "answer": "数学家"
}
<pinyin>duō duō yì shàn</pinyin>
输出：{
    "answer": "多多益善"
}
\end{verbatim}
本题目给出的拼音如下：\\
\verb|<pinyin>|\texttt{\textcolor{brown7}{\{pinyin\}}}\verb|</pinyin>|

\vspace{0.5cm}
\textbf{English Translation:}\\
You are a master of Chinese. Please provide the most commonly used phrase (including words or chengyu) corresponding to the given pinyin, and output the answer in JSON format. Example:
\begin{verbatim}
<pinyin>jiā fǎ jié hé lǜ</pinyin>
Output：{
    "answer": "加法结合律"
}
<pinyin>shù xué jiā</pinyin>
Output：{
    "answer": "数学家"
}
<pinyin>duō duō yì shàn</pinyin>
Output：{
    "answer": "多多益善"
}
\end{verbatim}
The pinyin given in this problem is as follows:\\
\verb|<pinyin>|\texttt{\textcolor{brown7}{\{pinyin\}}}\verb|</pinyin>|
\end{CJK}
\end{tcolorbox}
\caption{Prompt template for the pinyin2word test in RQ~1 (\S~\ref{app:RQ1}, for the Contextual Homophone Recognition task). \texttt{\textcolor{brown7}{\{pinyin\}}} is replaced with the actual pinyin.}
\label{fig:pmpt_hu_chr_py2phrase}
\end{figure*}

\begin{figure*}[ptb]
\centering
\begin{tcolorbox}

\textbf{Prompt Template}
\medskip\hrule\medskip

\begin{CJK}{UTF8}{gbsn}
\textbf{Chinese:}\\
你是一位使用汉语的作词人，需要根据既有的参考短句，创作出一句和它押韵的短句（且韵脚字不雷同）。注意，如果两个汉字的韵母出现在下面的同一组中，我们认为它们相互押韵（不要求声调相同）。
\begin{verbatim}
<groups>
第一组：a、ia、ua（如“他”“家”“瓜”）；
第二组：o、uo（如“波”“诺”）;
第三组：e（如“鹅”）；
第四组：ie、üe（如“接”“月”）；
第五组：er（如“而”）；
第六组：i、ü（如“知”“及”“鱼”）；
第七组：u（如“姑”）；
第八组：ai、uai（如“来”“乖”）；
第九组：ei、uei（如“雷”“鬼”）；
第十组：ao、iao（如“老”“交”）；
第十一组：ou、iou（如“楼”“有”）；
第十二组：an、uan、ian、üan（如“安”“关”“先”“远”）；
第十三组：en、uen（如“恩”“文”）；
第十四组：in、ün（如“音”“云”）；
第十五组：ang、uang、iang（如“昂”“光”“想”）；
第十六组：ing（如“英”）；
第十七组：eng、ueng（如"成"“风”“翁”）；
第十八组：ong、iong（如“龙”“雄”）。
</groups>
\end{verbatim}
请以json格式输出答案。示例如下：
\begin{verbatim}
<reference>一瞬间想起了那时候的自己</reference>
输出：{
"answer": "在塞纳河边的回忆你和谁在一起"
}
<reference>我有了你生命才有希望</reference>
输出：{
"answer": "破碎的心失去依傍"
}
<reference>付出所有的青春不留遗憾</reference>
输出：{
"answer": "你知道的我有半夜敲文字的习惯"
}
\end{verbatim}
请根据下面的参考短句生成押韵的短句：\\
\verb|<content>|\texttt{\textcolor{brown7}{\{content\}}}\verb|</content>|
\end{CJK} \\
\begin{flushright}
\textcolor{gray6}{\emph{(continue on next page)}}
\end{flushright}
\end{tcolorbox}
\end{figure*}
\begin{figure*}
\begin{tcolorbox}
\begin{CJK}{UTF8}{gbsn}
\textcolor{gray6}{\emph{(continue from previous page)}} \\[0.5cm]
\textbf{English Translation:}\\
You are a lyricist writing in Chinese. Based on the given reference sentence, create a new sentence that rhymes with it (the final character should not be the same as the reference's). Note: If the finals (yunmu) of two Chinese characters appear in the same group below, they are considered to rhyme (tone does not need to match).
\begin{verbatim}
<groups>
Group 1: a、ia、ua (e.g., "他", "家", "瓜");
Group 2: o、uo (e.g., "波", "诺");
Group 3: e (e.g., "鹅");
Group 4: ie、üe (e.g., "接", "月");
Group 5: er (e.g., "而");
Group 6: i、ü (e.g., "知", "及", "鱼");
Group 7: u (e.g., "姑");
Group 8: ai、uai (e.g., "来", "乖");
Group 9: ei、uei (e.g., "雷", "鬼");
Group 10: ao、iao (e.g., "老", "交");
Group 11: ou、iou (e.g., "楼", "有");
Group 12: an、uan、ian、üan (e.g., "安", "关", "先", "远");
Group 13: en、uen (e.g., "恩", "文");
Group 14: in、ün (e.g., "音", "云");
Group 15: ang、uang、iang (e.g., "昂", "光", "想");
Group 16: ing (e.g., "英");
Group 17: eng、ueng (e.g., "成", "风", "翁");
Group 18: ong、iong (e.g., "龙", "雄").
</groups>
\end{verbatim}
Please output the answer in JSON format. Example:
\begin{verbatim}
<reference>一瞬间想起了那时候的自己</reference>
Output: {
"answer": "在塞纳河边的回忆你和谁在一起"
}
<reference>我有了你生命才有希望</reference>
Output: {
"answer": "破碎的心失去依傍"
}
<reference>付出所有的青春不留遗憾</reference>
Output: {
"answer": "你知道的我有半夜敲文字的习惯"
}
\end{verbatim}
Please generate a sentence based rhyming with the reference sentence given below:\\
\verb|<reference>|\texttt{\textcolor{brown7}{\{content\}}}\verb|</reference>|
\end{CJK}
\end{tcolorbox}
\caption{Prompt template for the Rhyming Sentence Generation task (1-rhyme setting, \S~\ref{ssec:D2}). \texttt{\textcolor{brown7}{\{content\}}} is replaced with the actual reference sentence.}
\label{fig:pmpt_ru_r1}
\end{figure*}
\begin{figure*}[ptb]
\centering
\begin{tcolorbox}

\textbf{Prompt Template}
\medskip\hrule\medskip

\begin{CJK}{UTF8}{gbsn}
\textbf{Chinese:}\\
你是一位使用汉语的作词人，需要根据既有的参考短句，创作出一句和它双押的短句，要求它最后两个字和参考短句的最后两个字分别押韵（且韵脚字不雷同）。注意，如果两个汉字的韵母出现在下面的同一组中，我们认为它们相互押韵（不要求声调相同）。
\begin{verbatim}
<groups>
第一组：a、ia、ua（如“他”“家”“瓜”）；
第二组：o、uo（如“波”“诺”）;
第三组：e（如“鹅”）；
第四组：ie、üe（如“接”“月”）；
第五组：er（如“而”）；
第六组：i、ü（如“知”“及”“鱼”）；
第七组：u（如“姑”）；
第八组：ai、uai（如“来”“乖”）；
第九组：ei、uei（如“雷”“鬼”）；
第十组：ao、iao（如“老”“交”）；
第十一组：ou、iou（如“楼”“有”）；
第十二组：an、uan、ian、üan（如“安”“关”“先”“远”）；
第十三组：en、uen（如“恩”“文”）；
第十四组：in、ün（如“音”“云”）；
第十五组：ang、uang、iang（如“昂”“光”“想”）；
第十六组：ing（如“英”）；
第十七组：eng、ueng（如"成"“风”“翁”）；
第十八组：ong、iong（如“龙”“雄”）。
</groups>
\end{verbatim}
请以json格式输出答案。示例如下：
\begin{verbatim}
<reference>一瞬间想起了那时候的自己</reference>
输出：{
"answer": "孤独的我在梦里寻觅"
}
<reference>我有了你生命才有希望</reference>
输出：{
"answer": "多希望有一个让我依靠的肩膀"
}
<reference>付出所有的青春不留遗憾</reference>
输出：{
"answer": "用慈爱做花环，人与人不冷淡"
}
\end{verbatim}
请根据下面的参考短句生成押韵的短句：\\
\verb|<content>|\texttt{\textcolor{brown7}{\{content\}}}\verb|</content>| 
\end{CJK} \\
\begin{flushright}
\textcolor{gray6}{\emph{(continue on next page)}}
\end{flushright}
\end{tcolorbox}
\end{figure*}
\begin{figure*}
\begin{tcolorbox}
\begin{CJK}{UTF8}{gbsn}
\textcolor{gray6}{\emph{(continue from previous page)}} \\[0.5cm]
\textbf{English Translation:}\\
You are a lyricist writing in Chinese. Based on the given reference sentence, create a new double-rhymed sentence, such that the last two characters of your sentence rhyme with the last two characters of the reference sentence, respectively (the last two characters should not be the same as the reference's, too). Note: If the finals (yunmu) of two Chinese characters appear in the same group below, they are considered to rhyme (tone does not need to match).
\begin{verbatim}
<groups>
Group 1: a、ia、ua (e.g., "他", "家", "瓜");
Group 2: o、uo (e.g., "波", "诺");
Group 3: e (e.g., "鹅");
Group 4: ie、üe (e.g., "接", "月");
Group 5: er (e.g., "而");
Group 6: i、ü (e.g., "知", "及", "鱼");
Group 7: u (e.g., "姑");
Group 8: ai、uai (e.g., "来", "乖");
Group 9: ei、uei (e.g., "雷", "鬼");
Group 10: ao、iao (e.g., "老", "交");
Group 11: ou、iou (e.g., "楼", "有");
Group 12: an、uan、ian、üan (e.g., "安", "关", "先", "远");
Group 13: en、uen (e.g., "恩", "文");
Group 14: in、ün (e.g., "音", "云");
Group 15: ang、uang、iang (e.g., "昂", "光", "想");
Group 16: ing (e.g., "英");
Group 17: eng、ueng (e.g., "成", "风", "翁");
Group 18: ong、iong (e.g., "龙", "雄").
</groups>
\end{verbatim}
Please output the answer in JSON format. Example:
\begin{verbatim}
<reference>一瞬间想起了那时候的自己</reference>
输出：{
"answer": "在塞纳河边的回忆你和谁在一起"
}
<reference>我有了你生命才有希望</reference>
输出：{
"answer": "破碎的心失去依傍"
}
<reference>付出所有的青春不留遗憾</reference>
输出：{
"answer": "你知道的我有半夜敲文字的习惯"
}
\end{verbatim}
Please generate a sentence based double-rhyming with the reference sentence given below:\\
\verb|<reference>|\texttt{\textcolor{brown7}{\{content\}}}\verb|</reference>|
\end{CJK}
\end{tcolorbox}
\caption{Prompt template for the Rhyming Sentence Generation task (2-rhyme setting, \S~\ref{ssec:D2}). \texttt{\textcolor{brown7}{\{content\}}} is replaced with the actual reference sentence.}
\label{fig:pmpt_ru_r2}
\end{figure*}

\begin{figure*}[ptb]
\centering
\begin{tcolorbox}

\textbf{Prompt Template}
\medskip\hrule\medskip

\begin{CJK}{UTF8}{gbsn}
\textbf{Chinese:}\\
你是一位精通汉语的大师。请从四个选项中，选出在读音上和给出的词语最接近的一项（不要考虑语义），以json格式输出答案。示例如下：
\begin{verbatim}
<content>技法</content>
<choices>
A. 家法
B. 依法
C. 办法
D. 立法
</choices>
输出：{
    "answer": "D"
}
<content>增添</content>
<choices>
A. 今天
B. 春天
C. 昨天
D. 成天
</choices>
输出：{
    "answer": "B"
}
<content>渐渐</content>
<choices>
A. 健全
B. 建议
C. 间接
D. 见面
</choices>
输出：{
    "answer": "D"
}
\end{verbatim}
本题目给出的词语和选项如下：\\
\verb|<content>|\texttt{\textcolor{brown7}{\{content\}}}\verb|</content>|\\
\verb|<choices>|\\
\texttt{A. \textcolor{violet4}{\{A\}}}\\
\texttt{B. \textcolor{violet4}{\{B\}}}\\
\texttt{C. \textcolor{violet4}{\{C\}}}\\
\texttt{D. \textcolor{violet4}{\{D\}}}\\
\verb|</choices>|\\
\end{CJK}
\begin{flushright}
\textcolor{gray6}{\emph{(continue in the next page)}}
\end{flushright}
\end{tcolorbox}
\end{figure*}
\begin{figure*}
\begin{tcolorbox}
\begin{CJK}{UTF8}{gbsn}
\textcolor{gray6}{\emph{(continue from the previous page)}} \\[0.5cm]
\textbf{English Translation:}\\
You are a master of Chinese. From the four options, choose the one whose pronunciation is closest to the given word (ignore meaning), and output the answer in JSON format. Example:
\begin{verbatim}
<content>技法</content>
<choices>
A. 家法
B. 依法
C. 办法
D. 立法
</choices>
Output：{
    "answer": "D"
}
<content>增添</content>
<choices>
A. 今天
B. 春天
C. 昨天
D. 成天
</choices>
Output：{
    "answer": "B"
}
<content>渐渐</content>
<choices>
A. 健全
B. 建议
C. 间接
D. 见面
</choices>
Output：{
    "answer": "D"
}
\end{verbatim}
The word and choices given in this problem are as follows:\\
\verb|<content>|\texttt{\textcolor{brown7}{\{content\}}}\verb|</content>|\\
\verb|<choices>|\\
\texttt{A. \textcolor{violet4}{\{A\}}}\\
\texttt{B. \textcolor{violet4}{\{B\}}}\\
\texttt{C. \textcolor{violet4}{\{C\}}}\\
\texttt{D. \textcolor{violet4}{\{D\}}}\\
\verb|</choices>|\\
\end{CJK}
\end{tcolorbox}
\caption{Prompt template for the Similarity Comparison task (in word form, \S~\ref{ssec:D3}). \texttt{\textcolor{brown7}{\{content\}}} is replaced with the actual reference word, and \texttt{\textcolor{violet4}{\{A-D\}}} with the actual answer choices.}
\label{fig:pmpt_psu_sc_word4word}
\end{figure*}
\begin{figure*}[ptb]
\centering
\begin{tcolorbox}

\textbf{Prompt Template}
\medskip\hrule\medskip

\begin{CJK}{UTF8}{gbsn}
\textbf{Chinese:}\\
你是一位精通汉语的大师。请从四个拼音选项中，选出和给出的拼音读音最接近的一项，以json格式输出答案。示例如下：
\begin{verbatim}
<content>jì fǎ</content>
<choices>
A. jiā fǎ
B. yī fǎ
C. bàn fǎ
D. lì fǎ
</choices>
输出：{
    "answer": "D"
}
<content>zēng tiān</content>
<choices>
A. jīn tiān
B. chūn tiān
C. zuó tiān
D. chéng tiān
</choices>
输出：{
    "answer": "B"
}
<content>jiàn jiàn</content>
<choices>
A. jiàn quán
B. jiàn yì
C. jiàn jiē
D. jiàn miàn
</choices>
输出：{
    "answer": "D"
}
\end{verbatim}
本题目给出的拼音和选项如下：\\
\verb|<content>|\texttt{\textcolor{brown7}{\{content\}}}\verb|</content>|\\
\verb|<choices>|\\
\texttt{A. \textcolor{violet4}{\{A\}}}\\
\texttt{B. \textcolor{violet4}{\{B\}}}\\
\texttt{C. \textcolor{violet4}{\{C\}}}\\
\texttt{D. \textcolor{violet4}{\{D\}}}\\
\verb|</choices>|\\
\end{CJK}
\begin{flushright}
\textcolor{gray6}{\emph{(continue in the next page)}}
\end{flushright}
\end{tcolorbox}
\end{figure*}
\begin{figure*}
\begin{tcolorbox}
\begin{CJK}{UTF8}{gbsn}
\textcolor{gray6}{\emph{(continue from the previous page)}} \\[0.5cm]
\textbf{English Translation:}\\
You are a master of Chinese. From the four pinyin options, choose the one whose pronunciation is closest to the given one, and output the answer in JSON format. Example:
\begin{verbatim}
<content>jì fǎ</content>
<choices>
A. jiā fǎ
B. yī fǎ
C. bàn fǎ
D. lì fǎ
</choices>
Output：{
    "answer": "D"
}
<content>zēng tiān</content>
<choices>
A. jīn tiān
B. chūn tiān
C. zuó tiān
D. chéng tiān
</choices>
Output：{
    "answer": "B"
}
<content>jiàn jiàn</content>
<choices>
A. jiàn quán
B. jiàn yì
C. jiàn jiē
D. jiàn miàn
</choices>
Output：{
    "answer": "D"
}
\end{verbatim}
The pinyin and choices given in this problem are as follows:\\
\verb|<content>|\texttt{\textcolor{brown7}{\{content\}}}\verb|</content>|\\
\verb|<choices>|\\
\texttt{A. \textcolor{violet4}{\{A\}}}\\
\texttt{B. \textcolor{violet4}{\{B\}}}\\
\texttt{C. \textcolor{violet4}{\{C\}}}\\
\texttt{D. \textcolor{violet4}{\{D\}}}\\
\verb|</choices>|\\
\end{CJK}
\end{tcolorbox}
\caption{Prompt template for the Similarity Comparison task (in pinyin form, \S~\ref{ssec:D3}). \texttt{\textcolor{brown7}{\{content\}}} is replaced with the actual reference pinyin, and \texttt{\textcolor{violet4}{\{A-D\}}} with the actual answer choices.}
\label{fig:pmpt_psu_sc_py4py}
\end{figure*}

\begin{figure*}[ptb]
\centering
\begin{tcolorbox}

\textbf{Speech Template}
\medskip\hrule\medskip

\begin{CJK}{UTF8}{gbsn}
\textbf{Chinese:}\\
你是一位精通汉语的大师。请听写出下面我读出的四字成语，以json格式输出答案。本题给出的成语是:\texttt{\textcolor{brown7}{\{content\}}}。

\vspace{0.5cm}
\textbf{English Translation:}\\
You are a master of Chinese. Please transcribe the following four-character chengyu I read aloud and output the answer in JSON format. The given chengyu is: \texttt{\textcolor{brown7}{\{content\}}}.
\end{CJK}
\end{tcolorbox}
\caption{Speech template used for text-to-speech in the audio-model evaluation (Homophone Recall). \texttt{\textcolor{brown7}{\{content\}}} is replaced with the actual chengyu.}
\label{fig:speech_speech2idiom}
\end{figure*}

\begin{figure*}[ptb]
\centering
\begin{tcolorbox}

\textbf{Prompt Template}
\medskip\hrule\medskip

\begin{CJK}{UTF8}{gbsn}
\textbf{Chinese:}\\
请回答音频中的问题。\\
\texttt{\textcolor{olive6}{\{audio\_{file}\}}}

\vspace{0.5cm}
\textbf{English Translation:}\\
Please answer the question in the audio.\\
\texttt{\textcolor{olive6}{\{audio\_{file}\}}}
\end{CJK}
\end{tcolorbox}
\caption{Prompt template for in the audio-model experiment (\S~\ref{sssec:Extended Experiment}), paired with the audio file generated via text-to-speech. \texttt{\textcolor{olive6}{\{audio\_{file}\}}} is replaced with the actual audio file.}
\label{fig:pmpt_speech}
\end{figure*}

\begin{figure*}[ptb]
\centering
\begin{tcolorbox}
\textbf{Human Instruction}
\medskip\hrule\medskip
\begin{CJK}{UTF8}{gbsn}
\textbf{Chinese:}\\
任务：\\
写出对应汉字的普通话拼音。声调用数字附在最后（即第几声）。如：\\
问题：一\\
回答：yi1\\
\\
不会的可以猜一个。\\
闭卷！

\vspace{0.5cm}
\textbf{English Translation:}\\
Write the Mandarin pinyin corresponding to each given Chinese character. Indicate the tone by appending a number at the end (representing which tone it is). For example:\\
Question: 一\\
Answer: yi1\\
\\
If you’re unsure, you may make a guess.\\
Closed-book!
\end{CJK}
\end{tcolorbox}
\caption{Instruction for the character-to-pinyin verification experiment (\S~\ref{ssec:Verification experiment}).}
\label{fig:inst_verify_char2py}
\end{figure*}
\begin{figure*}[ptb]
\centering
\begin{tcolorbox}
\textbf{Human Instruction}
\medskip\hrule\medskip
\begin{CJK}{UTF8}{gbsn}
\textbf{Chinese:}\\
任务：\\
猜一个和给出的``短语''读音上最接近的成语。\\
由于拼音输入法本身就可能泄露答案，所以应当先猜测完再输入。（不能根据输入法的提示修改）。\\
如果猜不到，则在空格中填一个句号。

\vspace{0.5cm}
\textbf{English Translation:}\\
Task:\\
Guess the idiom whose pronunciation is closest to the given ``phrase.''\\
Since pinyin input methods may inadvertently reveal the answer, you should make your guess before typing it (do not rely on input method suggestions).\\
If you cannot guess, write a period (.) in the blank.
\end{CJK}
\end{tcolorbox}
\caption{Instruction for the Homophone Recall task (\S~\ref{sssec:Homophone Recall}).}
\label{fig:inst_hu_hc_homo2idiom}
\end{figure*}
\begin{figure*}[ptb]
\centering
\begin{tcolorbox}
\textbf{Human Instruction}
\medskip\hrule\medskip
\begin{CJK}{UTF8}{gbsn}
\textbf{Chinese:}\\
任务：\\
从这些谐音梗句子中，提取它们的谐音部分，以及谐音部分所对应的常用语。如：\\
句子：大户人家的小王犯了两项家规，父亲决定对小王数罪并罚，也就是家法结合律。\\
谐音部分：家法结合律\\
对应常用语：加法结合律\\
\\
注意，选取的时候尽可能地往小范围取，而非往大范围取。比如上面的例子中，谐音部分、对应常用语可以分别只写写``家法''和``加法''。

\vspace{0.5cm}
\textbf{English Translation:}\\
Task:\\
From these homophonic pun sentences, extract the homophonic part and the common phrase it corresponds to. For example:\\
Sentence: 大户人家的小王犯了两项家规，父亲决定对小王数罪并罚，也就是家法结合律。\\
Homophonic part: 家法结合律\\
Corresponding common phrase: 加法结合律\\
\\
Note: When selecting the phrase, try to keep the scope as narrow as possible rather than overly broad. For instance, in the example above, the homophonic part and the corresponding phrase could simply be ``家法'' and ``加法'', respectively.
\end{CJK}
\end{tcolorbox}
\caption{Instruction for the Contextual Homophone Recognition task (\S~\ref{sssec:Contextual Homophone Recognition}).}
\label{fig:inst_hu_chr_pun}
\end{figure*}
\begin{figure*}[ptb]
\centering
\begin{tcolorbox}
\textbf{Human Instruction}
\medskip\hrule\medskip
\begin{CJK}{UTF8}{gbsn}
\textbf{Chinese:}\\
任务：\\
从4个选项中，选择你感觉的和参考词读音最接近的一项（只考虑读音，不考虑语义）。\\
如果不确定，请蒙一个。

\vspace{0.5cm}
\textbf{English Translation:}\\
Task:\\
From the four options, choose the one whose pronunciation is closest to the reference word (consider sound only, not meaning).\\
If you are unsure, make a guess.
\end{CJK}
\end{tcolorbox}
\caption{Instruction for the Similarity Comparison task (in word form, \S~\ref{ssec:D3}).}
\label{fig:inst_psu_sc_word4word}
\end{figure*}
\begin{figure*}[ptb]
\centering
\begin{tcolorbox}
\textbf{Human Instruction}
\medskip\hrule\medskip
\begin{CJK}{UTF8}{gbsn}
\textbf{Chinese:}\\
任务：\\
从4个选项中，选择你感觉的和参考词读音最接近的一项。\\
如果不确定，请蒙一个。

\vspace{0.5cm}
\textbf{English Translation:}\\
Task:\\
From the four options, choose the one whose pronunciation you think is closest to the reference word.\\
If you are unsure, make a guess.
\end{CJK}
\end{tcolorbox}
\caption{Instruction for the Similarity Comparison task (in pinyin form, \S~\ref{ssec:D3}).}
\label{fig:inst_psu_sc_py4py}
\end{figure*}

\end{document}